\documentclass[sigconf]{acmart}

\usepackage{amsmath}
\usepackage{multirow}

\usepackage{hyperref}
\usepackage{amsfonts}
\usepackage{graphicx}
\usepackage[labelfont=normalfont]{subcaption}
\usepackage{float}
\usepackage{booktabs}
\usepackage{mathtools}

\usepackage{caption}

\newcommand{\GConv}{\ast_\mathcal{G}}
\usepackage{siunitx}
\usepackage{dsfont}
\newcount\Comments  
\usepackage{color}
\definecolor{darkgreen}{rgb}{0,0.5,0}
\definecolor{purple}{rgb}{1,0,1}
\newcommand{\kibitz}[2]{\ifnum\Comments=0\textcolor{#1}{#2}\fi}

\newcommand{\edit}[1]{\kibitz{black} {#1}}

\AtBeginDocument{%
  \providecommand\BibTeX{{%
    \normalfont B\kern-0.5em{\scshape i\kern-0.25em b}\kern-0.8em\TeX}}}

\copyrightyear{2023}
\acmYear{2023}
\setcopyright{acmlicensed}\acmConference[WWW '23]{Proceedings of the ACM Web Conference 2023}{May 1--5, 2023}{Austin, TX, USA}
\acmBooktitle{Proceedings of the ACM Web Conference 2023 (WWW '23), May 1--5, 2023, Austin, TX, USA}
\acmPrice{15.00}
\acmDOI{10.1145/3543507.3583452}
\acmISBN{978-1-4503-9416-1/23/04}

\begin{document}

\title{Detecting Socially Abnormal Highway Driving Behaviors via Recurrent Graph Attention Networks}


\author{Yue Hu, Yuhang Zhang, Yanbing Wang, Daniel Work}
\affiliation{%
  \institution{Vanderbilt University}
  \city{Nashville}
  \country{US}
}
\email{{yue.hu, yuhang.zhang.1,yanbing.wang,dan.work}@vanderbilt.edu}



\begin{abstract}
With the rapid development of Internet of Things technologies, the next generation traffic monitoring infrastructures are connected via the web, to aid traffic data collection and intelligent traffic management. One of the most important tasks in traffic is anomaly detection, since abnormal drivers can reduce traffic efficiency and cause safety issues. This work focuses on detecting abnormal driving behaviors from trajectories produced by highway video surveillance systems. Most of the current abnormal driving behavior detection methods focus on a limited category of abnormal behaviors that deal with a single vehicle without considering vehicular interactions. In this work, we consider the problem of detecting a variety of socially abnormal driving behaviors, i.e., behaviors that do not conform to the behavior of other nearby drivers.
This task is complicated by the variety of vehicular interactions and the spatial-temporal varying nature of highway traffic. To solve this problem, we propose an autoencoder with a Recurrent Graph Attention Network that can capture the highway driving behaviors contextualized on the surrounding cars, and detect anomalies that deviate from learned patterns. Our model is scalable to large freeways with thousands of cars. Experiments on data generated from traffic simulation software show that our model is the only one that can spot the exact vehicle conducting socially abnormal behaviors, among the state-of-the-art anomaly detection models. We further show the performance on real world HighD traffic dataset, where our model
detects vehicles that violate the local driving norms.

\end{abstract}




\maketitle

\section{Introduction}
\subsection{Motivation and challenges}
Nowadays, the development of Internet of Things (IoT) technologies has greatly advanced intelligent monitoring and management of urban transportation systems. Sensor networks including highway surveillance cameras and radar detectors, combined with Web of Things (WoT) technologies, have enabled transportation authorities to intelligently monitor traffic systems at scales and resolutions previously out of reach~\cite{trirat2021df,motionSystem,tang2019joint}.
One of the emerging tasks in intelligent traffic management is anomaly detection, since abnormal drivers could have adversarial impact on the smoothness of traffic stream, or even pose safety concerns. Yet it is infeasible for human operators to manually inspect and analyze all of this data, given the now massive amount of data these systems can generate. Consequently, there is need to spot the anomalies from terabytes of data and highlight the scenes that need further human inspection. In this work, we tackle the task of detecting abnormal driving behaviors, from trajectories produced by IoT highway video surveillance systems.

The existing approaches to vehicular anomaly detection mainly fall into two categories. The first set of approaches~\cite{bai2019traffic,chen2021dual,wu2021box,sultani2018real,li2020multi} focus on detecting severe events that cause vehicles to stop, and turn the problem into detecting stalled cars via computer vision from surveillance videos. The second set of approaches~\cite{matousek2019detecting,moukafih2019aggressive,alkinani2020detecting,gatteschi2021comparing} focus on single-car abnormal driving behaviors, such as speeding and abrupt braking. Methods ranging from thresholding~\cite{fazeen2012safe,carlos2018evaluation} to machine learning~\cite{matousek2018robust,matousek2019detecting,moukafih2019aggressive} are applied on data obtained from a single car, e.g. from on-board sensors and GPS devices. 

However, the above approaches only cover a subset of abnormal driving behaviors and treat cars in isolation from each other. Vehicles constantly interact with their surroundings, and traffic context is needed for anomaly detection. For example, a car at a constant speed of 50 mph might be perfectly normal, yet a car at 50 mph on the inner most lane of highway is blocking all the other cars driving at 65 mph or above, and should be considered an anomaly on highway. As another example, abrupt braking might be considered abnormal, but if the vehicle is braking because its front car is stalled, then we should detect the front car as abnormal, whereas the braking car is doing what is expected. Our task is to detect such \textit{socially abnormal behaviors}  that do not conform to the commonly accepted and observed social norms, by developing a contextual understanding of vehicle interactions. 

Moreover, most of the existing methods are rule-based~\cite{fazeen2012safe,carlos2018evaluation} or supervised-learning method~\cite{moukafih2019aggressive,gatteschi2021comparing}, and can only detect pre-defined types of anomalies such as stalled and speeding cars. Yet vehicles can behave anomalously in unexpected ways, and building a comprehensive set of rules that includes every possible occasion could be hard. Our goal is to build a model that can identify a variety of anomalies with unsupervised learning.

Graph neural networks (GNN) have seen rapid development in recent years~\cite{kipf2016semi,velivckovic2017graph,wang2022graph,wang2021tree}, showing great advantage in modeling the complex relationships in graph data. By representing vehicles as nodes and their relationships as edges, interactions with neighbors can be modeled via a GNN. 
Two challenges exist in developing a GNN for detecting anomalous driving behaviors. First, given the spatial-temporally varying nature of trajectories, we need to deal with dynamic graphs. This is a nontrivial problem as compared to anomaly detection on only a static graph, or only considering time-varying signals. Second, we need to take stochasticity into consideration, which is intrinsic in driving behaviors. Under a particular context, there could be a range of acceptable behaviors that should all be considered normal. For example, a car can conduct lane changing from time to time, or have some variation in speed. A deterministic model that fails to capture such normal stochasticity would mislabel every lane-changing car as an anomaly.

\subsection{Our approach}

To detect socially abnormal behaviors on large scale trajectory data while addressing the aforementioned challenges, we develop a model for \textit{Detecting Socially Abnormal Behaviors} \textbf{(DSAB)} in highway driving via Recurrent Graph Attention Autoencoder. Graph attention networks combined with recurrent neural networks is used to capture the spatial-temporal pattern of vehicle trajectories,  while dynamically taking each vehicle's neighbors into consideration based on vehicle states. To facilitate scalability while capturing anomalous driving behaviors occurring over longer periods of time, we sample the trajectories over a relatively long-horizon and coarse-grid time window. We further use a sparse graph where the vehicles are only connected to close neighbors to reduce computation. An autoencoder structure is used for anomaly detection, which can encode and decode normal data well. To address the stochasticity in driving behaviors, we reconstruct the probabilistic distribution of the trajectories in decoding process. Samples with small reconstruction probabilities are marked as anomalies.


With these designs, our model can detect the socialy abnormal driving behaviors, and is scalable to thousands of cars over 5 miles of highway. We show the effectiveness of our model on both simulation and real-world data. First, we generate large-scale trajectory data with ground-truth anomaly labels, via a microscopic traffic simulator, to quantitatively evaluate the performance. Compared with the existing state-of-the-art methods, our model is the only one that can detect the exact vehicle with abnormal behavior, whereas the rest models can only detect anomalous scenes
We also apply our method on the real-world highD trajectory dataset~\cite{highDdataset}, where our model detects vehicles that violate the local driving norms.

To summarize, the contribution of our works is as follows:

\begin{itemize}
    \item We propose a new problem of detecting the exact anomalous vehicles that violate the social interaction norms in highway driving, and develop a model to solve it, achieving state of the art performance. 
    \item We develop a DSAB model based on Recurrent Graph Attention Networks. It well captures the spatial-temporal trajectory dynamics, while considering both the vehicular interactions and the stocasticity in driving behaviors. 
    \item We conduct extensive experiments on both simulation and real-world data sets, and show the ability of our model to scale to large highway monitoring systems with thousands of vehicles, and detect a variety of abnormal behaviors.

\end{itemize}

\section{Related work} \label{Sec:related}
\noindent\textbf{Trajectory Modeling.}
The majority of trajectory modeling work focuses on the prediction of future trajectories for humans or vehicles, and in this line, modeling the interaction between the agents is gaining interest. To aggregate information across agents, a pooling mechanism is used in the Social LSTM~\cite{alahi2016social} and the Social GAN~\cite{gupta2018social}, while an attention mechanism is used in SoPhie~\cite{sadeghian2019sophie}, and scene context fusion via convolutional neural network is used in Desire~\cite{lee2017desire}. With recent development of graph convolutional networks, works including \cite{mohamed2020social,vemula2018social,huang2019stgat} model the agents as nodes and their relationships as edges, and develop spatio-temporal GNN to learn the dynamics.

\noindent\textbf{Anomaly detection.}
\edit{Anomaly detection has been an important task in transportation. Unexpected autonomous driving condition detection is studied in~\cite{stocco2022thirdeye,stocco2020misbehaviour}. Extreme event detection in urban traffic is studied in~\cite{hu2020robust,hu2022detecting}.
Our work detects anomalies on graphs, and a comprehensive survey on graph anomaly detection can be found in~\cite{ma2021comprehensive}. Common detection methods on graphs include autoencoder-based methods~\cite{ding2019deep,bandyopadhyay2020outlier}, generative adversarial learning~\cite{ding2021inductive}, and contrastive learning~\cite{xu2022contrastive}. However, most of the existing works are on static graphs, whereas we need to deal with dynamic graphs, adding the temporal dimension. The existing methods for dynamic graph anomaly detection have different problem settings compared to ours. Specifically, NetWalk~\cite{yu2018netwalk} and TADDY~\cite{liu2021anomaly} deal with unattributed graphs with no node or edge features; AddGraph~\cite{zheng2019addgraph} and StrGNN~\cite{cai2021structural} detect anomalous edges.}

The most relevant work with ours is STGAE~\cite{wiederer2022anomaly}, where a spatio-temporal graph autoencoder is combined with  kernel density estimation (KDE) to detect abnormal driving behaviors. Yet while STGAE works well in experiments with only two vehicles, the time complexity of KDE is too high for detection in large numbers of vehicles. 
Moreover, STGAE tackles a different task of detecting the existence of anomalies among several vehicles over a stretch of road over a specific time, which we term as \textit{abnormal scene detection}.  Compared with our task of detecting the specific abnormal vehicles, abnormal scene detection is an easier task, since an abnormal vehicle can have subsequent influence on its neighbor vehicles. For example, a slow or stalled car can cause its following cars to break abruptly or change lanes, which could be detected as abnormal, but should have been considered normal when looking for the root cause. We demonstrate in our experiments that while most baseline methods can work well on abnormal scene detection, they achieve poor performance on abnormal vehicle detection.

\section{Method}
\begin{figure}
    \centering
    \includegraphics[width=\linewidth]{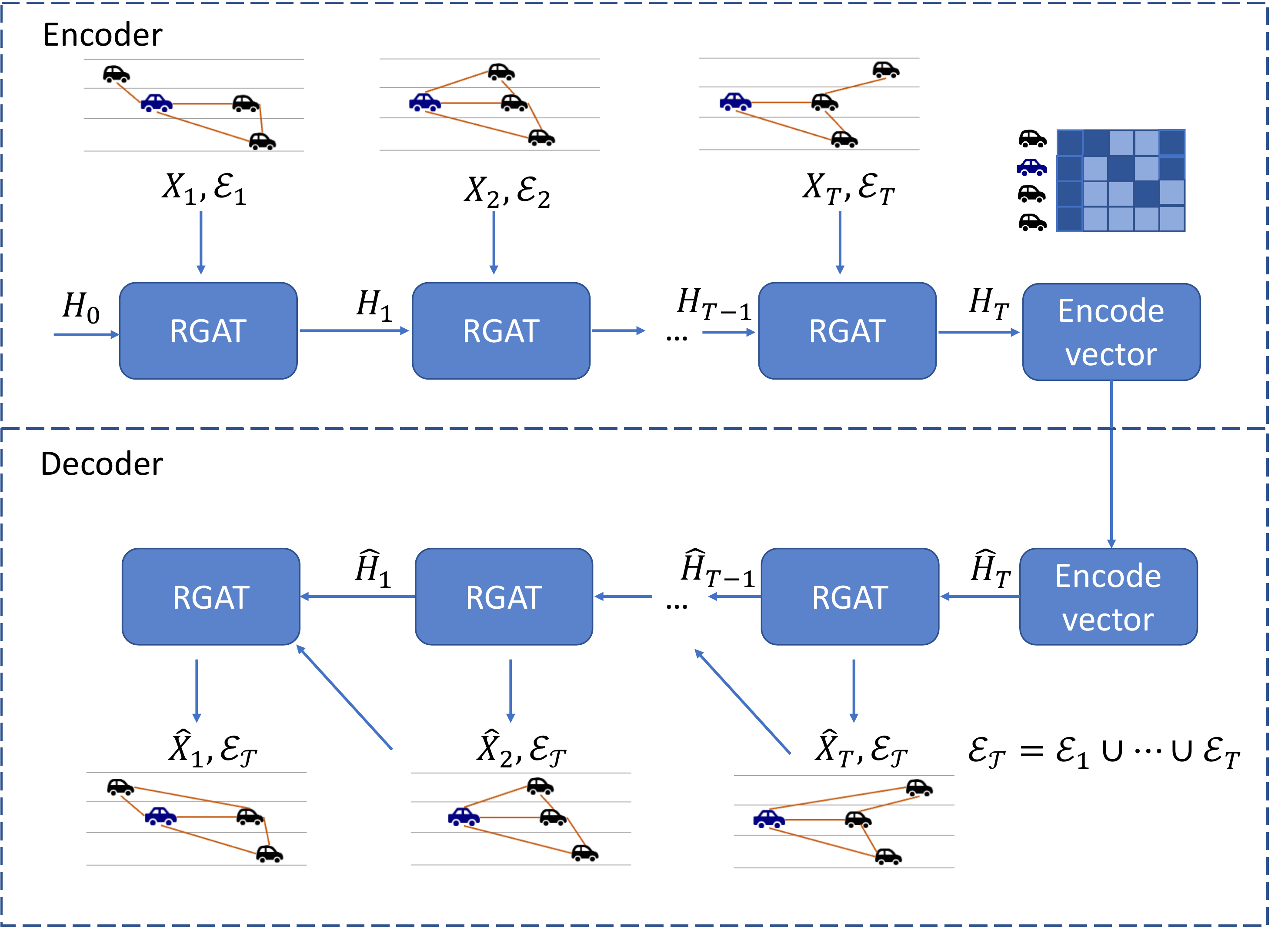}
    \caption{DSAB model overview. We construct a spatial-temporal dynamic graph to represent vehicles. An encoder compresses the dynamic graph into a low-dimensional encode vector, and a decoder reconstructs the input vehicle states, based on Recurrent Graph Attention Network (RGAT). In the encoder, $\mathbf{X}_t, \mathcal{E}_t$ denotes the node feature matrix and edge set respectively. In the decoder,  $\hat{\mathbf{X}_t}$ is the node reconstruction matrix, $ \mathcal{E}_\mathcal{T}$ is the union of all input edge sets.  
    }
    \label{fig:model}
    \Description{This figure has two parts. The top part shows the framework for the encoder where $\T+1$ RGAT is displayed with corresponding inputs and outputs. Similarly, the bottom part shows the framework for the decoder where $\T+1$ RGAT is displayed with corresponding inputs and outputs.}
\end{figure}

In this section, we formulate the problem mathematically, then describe the proposed DSAB model for anomaly detection on highways. In an overview, we first construct the vehicle trajectories as a spatial-temporal dynamic graph. Then we build an autoencoder with an encoder to compress the vehicle time series in low-dimensional vectors, and a decoder to reconstruct the probabilistic distribution of original input trajectories. At test time, we use the reconstruction probability as a measurement of anomalous behavior. The overview of the model is shown in Fig~\ref{fig:model}.

We further make the following considerations to best accommodate large scale anomaly detection on highway driving. Unlike most works~\cite{lee2017desire, alahi2016social,mohamed2020social} that works on short-term fine-grained trajectories, e.g., 3-10 Hz over 1-5s, we choose coarser-grained sampling with larger time window. In addition to reducing the amount of data needed to be processed, it also captures the anomalous driving behaviors that are expected to occur over longer periods of time. E.g., a vehicle that is speeding or tailgating will likely persist for more than a few seconds.  While computationally advantageous, sampling at a more coarse timescale requires a different model to be used for lane changing.  Specifically, the continuous bi-variate Gaussian distribution, which is commonly used in trajectory modeling~\cite{vemula2018social,mohamed2020social,alahi2016social}, is no longer suitable to capture a discrete lane-changing motion. Instead, we model longitudinal motions as Gaussian distribution, and lateral lane location as categorical distribution. 

\subsection{Problem formulation} \label{sec:formulation}

Our model input is the observations of a set of $N$ vehicles on a highway over time window $\mathcal{T}$, where $N$ can vary for different time windows. For each vehicle $i$ at time $t\in \mathcal{T}$, the observation $\mathbf{o}_t^i = [x_t^i, y_t^i, l_t^i, v_t^i, a_t^i]$ includes longitudinal position $x_t^i$, lateral position $y_t^i$, driving lane id $l_t^i$, longitudinal speed $v_t^i$, and longitudinal acceleration $a_t^i$. 

Given the vehicle observations $\mathbf{o}_t^i$, $\forall t \in \mathcal{T}$, our major goal is to detect the \textit{vehicles} that have abnormal behavior during the time window $\mathcal{T}$. Additionally, we also report the performance of detecting abnormal \textit{scenes} - dividing the entire highway into short stretches of length $\delta_s$, a scene contains all cars on a stretch $\mathcal{S}$ during $\mathcal{T}$, and is labeled as abnormal if it contains any abnormal cars.

\subsection{Graph construction for vehicle trajectories } \label{Sec:graph_const}
We first construct a dynamic graph for the vehicle trajectories during time window $\mathcal{T}$, denoted as $G(\mathcal{T}) = \{\left(\mathcal{V}, \mathcal{E}_1\right), \left(\mathcal{V}, \mathcal{E}_2\right), \ldots,$ $ \left(\mathcal{V}, \mathcal{E}_T\right)\}$ for discrete time steps $t \in \mathcal{T} = \{1,2, \dots, T \}$. The node set $\mathcal{V}$ includes all the vehicles on highway during the time window $\mathcal{T}$. The number of nodes $N=|\mathcal{V}|$ is fixed for the graph of a specific time window, but can vary for graphs of different time windows\footnote{We explain in implementation details in Section~\ref{sec:details} how we deal with vehicles that are not present in the highway during the entire time window.} At each time step $t$, the vehicle observations are summarized into a raw observation matrix $\mathbf{O}_t \in \mathbb{R}^{N \times d}$. $\mathbf{O}_t$ includes the observation $\mathbf{o}_t^i = [x_t^i, y_t^i, l_t^i, v_t^i, a_t^i]$ for all vehicles $V_i \in \mathcal{V}$.
There is an edge $e^{ij}_t \in \mathcal{E}_t$ if vehicle $i$ and $j$ are close at time $t$, that is, the two vehicles are less than $\delta_x$ feet apart longitudinally, and less than $\delta_l$ lanes apart laterally. This is based on the consideration that in reality, the vehicles are most influenced by their close neighborhood vehicles. Thus we limit the neighbor range to reduce storage and computational requirements. 

 
\subsection{Encoder} 
In this subsection, we explain how we use \textit{Recurrent Graph Attention Network} (RGAT) to encode the spatial-temporal vehicle trajectories. We adopt an RNN which has proven to work well on time series data. Furthermore, to consider the dynamic influence of each vehicle's neighbors, we integrate the graph attention network with RNN.

Specifically, we adopt the GRU~\cite{cho2014properties,chung2014empirical} variant of RNN, which is capable of learning long range dependencies in time via a gating mechanism. Furthermore, similar to the practice of~\cite{li2018diffusion,seo2018structured}, we substitute the matrix multiplications in the original GRU with graph convolutions denoted as $\GConv$, to operate on the input and hidden states and capture neighborhood interactions:
\begin{equation}
\label{eq:GNN_RNN}
\begin{aligned}
    & \mathbf{z}_{t} = \sigma (\mathbf{W}_{xz}\GConv \mathbf{X}_t + \mathbf{W}_{hz}\GConv \mathbf{H}_{t-1}+\mathbf{b}_z),\\
    & \mathbf{r}_{t} = \sigma (\mathbf{W}_{xr}\GConv \mathbf{X}_t + \mathbf{W}_{hr}\GConv \mathbf{H}_{t-1}+\mathbf{b}_r),\\
    & \Tilde{\mathbf{H}}_{t} = \text{tanh}(\mathbf{W}_{xh}\GConv \mathbf{X}_t + \mathbf{W}_{hh}\GConv (\mathbf{r}_t\odot \mathbf{H}_{t-1} +\mathbf{b}_h)),\\
    & \mathbf{H}_t = \mathbf{z}_t \odot \mathbf{H}_{t-1} + (1-\mathbf{z}_t) \odot \Tilde{\mathbf{H}}_t,
    \end{aligned}
\end{equation}
where $\mathbf{H}_t  \in \mathbb{R}^{N \times d_h}$ is the hidden state,  $\mathbf{z}_t \in \mathbb{R}^{N \times d_h}$ and  $\mathbf{r}_t \in \mathbb{R}^{N \times d_h}$ are the update gate and the reset gate respectively, with hidden dimension $d_h$. The weights $\mathbf{W}_{xz}$,  $\mathbf{W}_{hz}$,  $\mathbf{W}_{xr}$, $\mathbf{W}_{hr}$, $\mathbf{W}_{xh}$, $\mathbf{W}_{hh}$ and biases $\mathbf{b}_z$, $\mathbf{b}_r$, $\mathbf{b}_h$ are trainable parameters.  $\sigma$ is the sigmoid function, and  $\odot$ is element-wise multiplication. $\mathbf{X}_t \in \mathbb{R}^{N \times d_f}$ is the input at time step $t$, derived from the raw observation input $\mathbf{O}_t$, and consists of numerical observations of position, speed, and acceleration $[x_t , y_t, v_t, a_t] \in \mathbb{R}^{N \times 4}$ , concatenated with lane embeddings $ \mathbf{h}_{\text{lane}_t}\in \mathbb{R}^{N \times d_l}$ for categorical observations of driving lane IDs $\mathbf{l}_t$. Thus, the input feature dimension $d_f = 4+d_l$. Each driving lane ID is mapped to a corresponding embedding vector with dimension $d_l$. The entries of the lane embedding vectors are initialized at random and learned during training.
 
In terms of the graph convolution operator $\GConv$, instead of the Chebyshev spectral graph convolutional operator~\cite{defferrard2016convolutional} adopted by~\cite{seo2018structured}, which uses pre-defined edge weights, we adopt graph convolutions based on graph attention mechanism~\cite{brody2021attentive,velivckovic2017graph}, which determines the relevance of the vehicle's neighbors dynamically based on the vehicle states. Next we introduce the graph attention based convolution $\mathbf{W} \GConv \mathbf{X}_t$ operated  on the input observation matrix $\mathbf{X}_t$, whereas $\mathbf{W} \GConv \mathbf{H}_t$ follows the same process on hidden states $\mathbf{H}_t$ with corresponding weights. Denoting $\mathbf{x}_t^i  \in \mathbb{R}^{d_f}$ as the entry for vehicle $i$ in $\mathbf{X}_t$,  $\mathcal{N}_t^i$ as the set of neighbors of node $i$ at $t$, $\mathbf{W} \GConv \mathbf{X}_t$ works as follows:
\begin{equation}
\label{eq:GAT}
    \mathbf{x}^{\prime i}_t = \mathbf{W} \GConv \mathbf{X}_t \coloneqq \alpha_{i,i}\mathbf{W}\mathbf{x}_{t}^i +
        \sum_{j \in \mathcal{N}_t^i} \alpha_{i,j}\mathbf{W}\mathbf{x}_{t}^j,
\end{equation}
where $\mathbf{x}^{\prime i}_t  \in \mathbb{R}^{d_h}$ is the output node embedding, $\mathbf{W} \in \mathbb{R}^{d_h \times d_f}$ is the weight matrix.  The term $\alpha_{i,j}$ is the attention score calculated as:

\begin{equation}
\label{eq:att}
\begin{aligned}
    & e_{i,j} = \left(\mathbf{a}^{\top}\mathrm{LeakyReLU}\left(
        [\mathbf{W}\mathbf{x}_t^i \, \Vert \, \mathbf{W}\mathbf{x}_t^j]
        \right)\right), \\
    & \alpha_{i,j} = \text{softmax}(e_{i,j}) 
        = \frac{\text{exp} (e_{i,j})}{\sum_{k \in \mathcal{N}_t^i}\text{exp} (e_{i,k})},
    \end{aligned}
\end{equation}
where $\Vert$ is the concatenate operator, and $\mathbf{a}\in\mathbb{R}^{2d_h} $ is a trainable weight vector. In~\eqref{eq:GAT} and \eqref{eq:att}, first every node input feature  goes through a linear transformation parameterized by $\mathbf{W}$. Then, the attention coefficients $e_{i,j}$ are calculated by concatenating the transformed node features, followed by a non-linear activation (LeakyReLU), and a linear transformation with parameter $\mathbf{a}$. Then, for each node, the attention coefficients of all its neighbors are normalized by a softmax operator to reach the attention scores $\alpha_{i,j}$. Finally, the output node embedding is calculated as the linear combination of its neighbor transformed feature vectors, weighted by the attention scores. 

Compared with works~\cite{mohamed2020social,wiederer2022anomaly} that pre-define edge weights as a function of physical distances, the attention mechanism we use has more expressive power, and can comprehensively determine the neighbor relevance based on observation information. Furthermore, the real influence of vehicles in front of and behind an ego car is asymmetric (e.g., you must slow down immediately for slow-moving cars in front of you, but not for slow cars behind you), and the attention formulation in~\eqref{eq:att} can achieve such asymmetry. In comparison, when using a function of distance as edge weights~\cite{mohamed2020social,wiederer2022anomaly}, the same importance is assigned to cars that are close to a vehicle, regardless if they are in front or behind the ego vehicle.

Moreover, \textit{multi-head attention} is used to attend to different aspects of the neighborhood information, similar to~\cite{vaswani2017attention,velivckovic2017graph}. Specifically, $K$ independent heads is used  following~\eqref{eq:GAT}, and the final results of the heads are averaged:
\begin{equation}
\label{eq:GAT_K}
    \mathbf{x}^{\prime i}_t = \frac{1}{K} \sum_{k = 1}^K\left( \alpha_{i,i}^k\mathbf{W}^k\mathbf{x}_{t}^i +
        \sum_{j \in \mathcal{N}_t^i} \alpha_{i,j}^k\mathbf{W}^k\mathbf{x}_{t}^j \right).
\end{equation}

The hidden states for all vehicles are updated at every time step by~\eqref{eq:GNN_RNN}, \eqref{eq:GAT} and \eqref{eq:att}. The final hidden state $\mathbf{H}_T$ is used as the encoded vector embedding.

\subsection{Decoder}
The decoder works in the same way as the encoder, using the RGAT structure. To avoid the computation burden of re-calculating the edge set thresholded by the reconstructed vehicle position at every step, we use the union of all edge sets at encoding time steps, $\mathcal{E_T} = \mathcal{E}_1 \cup  \mathcal{E}_2 \cup \dots \cup \mathcal{E}_T$. The edge set only limits the range of neighbors each node attends to, while the importance of neighbors are calculated dynamically with attention mechanism. The decoded hidden state $\hat{\mathbf{H}}_t  \in \mathbb{R}^{N \times d_h}$ further goes through a fully connected layer to produce the output vehicle state reconstruction $\hat{\mathbf{X}}_t$. The details of  $\hat{\mathbf{X}}_t$ will be explained in Section~\ref{Sec:loss}. Then, the output $\hat{\mathbf{X}}_t$ is used as input for the next recurrent step.  We decode the time series from time $T$ backwards, since the encoder vector is most relevant to the states at time $T$, which is most recently encoded. The initial input into the decoder GAT-RNN is the vehicle states at time $T$.


\subsection{Loss function} \label{Sec:loss}
For each vehicle $i$ at time $t$, we assume its longitudinal position $x_t^i$, speed $v^i_t$, and acceleration  $a^i_t$ each follows an univariate Gaussian distribution. That is, $x^i_t \sim \mathcal{N}\left(\mu_{x_t^i}, \sigma_{x_t^i}\right)$, $v^i_t \sim \mathcal{N}\left(\mu_{v_t^i}, \sigma_{v_t^i}\right)$, $a^i_t \sim \mathcal{N}\left(\mu_{a_t^i}, \sigma_{a_t^i}\right)$. Further, we assume the lateral lane position  $l_t^i$ is a discrete choice among $L$ lanes, with an underlying categorical distribution $\left\{p_{1_t^i}, \dots, p_{L_t^i}\right\}$.
The Gaussian distribution and categorical distribution parameters are estimated by decoder output $\hat{\mathbf{X}} $. That is, $\hat{\mathbf{x}}^i_t = \left[ \hat{\mu}_{x_t^i}, \hat{\sigma}_{x_t^i},\hat{\mu}_{v_t^i}, \hat{\sigma}_{v_t^i},\hat{\mu}_{a_t^i}, \hat{\sigma}_{a_t^i} , \hat{p}_{1_t^i}, \dots, \hat{p}_{L_t^i} \right]$ estimates the mean and variance of position, speed, and acceleration, as well as the probability of being in each lane.

Denoting the estimated probability density function of position, speed, and acceleration as $q\left(x^i_t | \hat{\mu}_{x_t^i}, \hat{\sigma}_{x_t^i}\right)$, $q\left(v^i_t |  \hat{\mu}_{v_t^i}, \hat{\sigma}_{v_t^i}\right)$, and $q\left(a^i_t | \hat{\mu}_{a_t^i}, \hat{\sigma}_{a_t^i}\right)$ respectively, we aim to minimize the negative log-likelihoods as follows:

\begin{equation}
\label{eq:NLL}
\begin{aligned}
    & \mathcal{L}_{x_t^i} = - \text{log}\left(q\left(x^i_t | \hat{\mu}_{x_t^i}, \hat{\sigma}_{x_t^i}\right)\right),\\
    & \mathcal{L}_{v_t^i} = - \text{log}\left(q\left(v^i_t | \hat{\mu}_{v_t^i}, \hat{\sigma}_{v_t^i}\right)\right),\\
    & \mathcal{L}_{a_t^i} = - \text{log}\left(q\left(a^i_t | \hat{\mu}_{a_t^i}, \hat{\sigma}_{a_t^i}\right)\right).\\
    \end{aligned}
\end{equation}

As for lane classification, we aim to minimize the cross entropy loss as follows:

\begin{equation}
    \label{eq:CEL}
    \mathcal{L}_{l_t^i} = -\sum_{l=1}^L \mathds{1}_{l_t^i} \text{log}\left(\hat{p}_{l_t^i}\right),
\end{equation}
where $\mathds{1}_{l_t^i}= 1$ if vehicle $i$ is in lane $l$ at time t and 0 otherwise. The final loss is a weighted sum of the negative log-likelihood losses and the cross entropy loss across all agents and all times:

\begin{equation}
    \label{eq:loss}
\begin{aligned}
    & \mathcal{L}^i_t = \lambda_{x} \mathcal{L}_{x_t^i} + \lambda_v \mathcal{L}_{v_t^i} + \lambda_a \mathcal{L}_{a_t^i} + \lambda_l \mathcal{L}_{l_t^i}, \\
     &\mathcal{L} = \sum_{i=1}^N \sum_{t = 1}^T \mathcal{L}^i_t,
\end{aligned}
\end{equation}
The weights are set as $\lambda_{x} = 1$ , $\lambda_v = 1$ , $\lambda_a = 2$ and $\lambda_l = 2$  empirically.

\subsection{Anomaly detection} \label{sec:agg}

For abnormal vehicle detection, the anomaly score for vehicle $i$ over time window $\mathcal{T}$ is calculated by averaging the loss over all time steps: 
\begin{equation}
    \label{eq:car_score}
\begin{aligned}
     \alpha^{i}_{\mathcal{T}} =  \frac{1}{T} \sum_{t = 1}^T\mathcal{L}_t^i,
\end{aligned}
\end{equation}
where $T$ is the length of time window $\mathcal{T}$.
For abnormal scene detection, we aggregate the loss of all the vehicles that appear in the stretch $\mathcal{S}$ during $\mathcal{T}$ as the stretch anomaly score:
\begin{equation}
    \label{eq:stretch}
\begin{aligned}
     \alpha^{\mathcal{S}}_{\mathcal{T}} = \text{max}(\mathcal{L}_t^i), \forall(i,t) \text{ that } x^i_t \in \mathcal{S}.
\end{aligned}
\end{equation}
The maximum aggregation is used instead of the mean for abnormal scene detection, so that the score is more sensitive to the existence of anomalies, and not averaged out by normal vehicles. Appendix~\ref{sec:aggregator} conducts a detailed comparison between maximizing and averaging.

\section{Experiments}\label{sec:experiment}
In this section, we use two data sources to evaluate the performance of our method. First, simulation data is used to quantitatively compare the performance of our method with the baselines, where we have the ground truth anomaly labels. Second, highD dataset~\cite{highDdataset} is used to qualitatively show our method works on real-world trajectories. The code is publicly available on GitHub\footnote{\url{https://github.com/yuehu9/DSAB-Detecting-Socially-Abnormal-Drving-Behaviors}}. 
\subsection{Datasets} \label{sec:dataset}
The detailed information of the two datasets is as follows.
\subsubsection{Simulation data}
TransModeler is a microscopic traffic simulator that generates vehicle trajectories mimicking human driving behavior and interactions. In this work, we generate a set of recordings at 1Hz on a 5-mile stretch of a 4-lane highway. The recordings have different traffic flows and vehicle type distribution to include different traffic conditions and abnormal scenarios. Specifically, we include the following scenarios: \begin{itemize}
    \item Normal traffic. A standard car following model, \textit{Modified General Motors} ~\cite{ahmed1999modeling} is used, which has been demonstrated to correlate well with field traffic data. The desired speed of the vehicles follows a typical distribution found in real-world traffic, with around 5\% speeding vehicles and 5\% slow vehicles. We include both free flow and congested conditions with varying traffic demands.
    
    \item Speeding. The abnormal speeding cars drive at least 15 mph faster than the other vehicles when it is possible to do so.
    
    \item Slow. The abnormal slow cars drive at least 15 mph slower than the other vehicles when it is possible to do so.
    
    \item Tailgating. The tailgating vehicle's headway is less than 0.5s from the lead car. We simulate tailgating cars via a \textit{Constant Time Gap} car-following model, where drivers can keep a constant desired headway from the leading vehicle~\cite{1337326}.
    
    \item Stalled car. A vehicle randomly stops on the road for a period. 
    
    \item Comprehensive scenario. We include all the above abnormal scenarios, i.e., speeding, slow, tailgating and stalled vehicles are all present in a single scenario, to comprehensively evaluate model detection performance. 
\end{itemize}

We include both common anomalies  (e.g., speeding, tailgating and stalled vehicles~\cite{matousek2019detecting,alkinani2020detecting,moukafih2019aggressive,bai2019traffic}) as well as those that are not commonly studied (e.g., slow driving). Slow driving can create moving bottlenecks with adverse impacts on traffic, but are difficult to identify using existing single-vehicle approaches, demonstrating the importance of developing an interaction-based detector.

The detailed experimental settings and distributions can be found in the Appendix. We adjust the percentages in each abnormal scenario, so that the abnormal behaving cars consist of around 3\%-5\% of total cars, randomly distributed on the road. The actual anomaly rate has some variation across different recordings because of simulation randomness. 
A total of 180 min of normal traffic is used for training. The recordings of speeding, slow, tailgating, stalled car and comprehensive scenarios are used for testing, each lasting 10 min. The training data has 7,886 cars in total, producing 920,311 trajectories when segmented into 15s windows at 1s stride. The testing sets together have a total of 5,067 cars and 630,718 trajectories. Detailed statistics can be found in the Appendix. 

We note that our training dataset is not perfectly clean, but contains a small portion of abnormal data (e.g., speeding and slow cars). This is intentionally done to simulate the real-world situation where we may not have a clean labeled training set that is known to be free of anomalous vehicles. \edit{Experiments excluding training anomalies can be found in Appendix~\ref{sec:clean_training}, which shows similar results.}

\subsubsection{Real-world data}
The HighD dataset~\cite{highDdataset} contains high accuracy vehicle trajectories extracted from video recordings captured via unmanned aerial vehicles over various stretches of German highways, each stretching approximately 1300 feet long. We select a total of approximately 350 minutes of recordings on three-lane highways as our training data, which covers both light and heavy traffic conditions, with traffic flow varying from 1200 to 3600 vehicles/lane/hour. Then, we test on an unseen 15-min recording with a flow of 2300 vehicles/lane/hour. The training data has 42,106 cars in total, producing  459,187 trajectories when segmented into 10s windows at 1s stride. The testing sets together has 1,795 cars and 21,840 trajectories. For the HighD data, we do not have the ground truth anomaly labels. Thus, we qualitatively examine the top anomalies detected by our model.

\subsection{Baselines and Metrics}
\subsubsection{Baselines}
We compare our method with simple heuristic methods as well as state-of-the-art methods for anomaly detection on trajectory data. The baselines include:
\textit{i})  \textit{Linear temporal interpolation} (LTI) implemented by~\cite{wiederer2022anomaly} that uses a linear interpolation between the first and last position of the vehicle to reconstruct the trajectory;
\textit{ii}) \textit{Constant Velocity Model} (CVM) ~\cite{scholler2020constant} that  assumes constant speed as recorded at the first observation time step to reconstruct the trajectory;
\textit{iii}) \textit{Robust tensor Recovery}  
 (RTR)~\cite{hu2020robust} model that captures spatial-temporal correlations via low-rank tensor decomposition, and detects sparse outliers that deviates from the normal patterns;
\textit{iv}) \textit{Seq2Seq model}~\cite{sutskever2014sequence} that encodes and decodes the time series with two LSTM networks, and uses \textit{Mean Square Error} (MSE) as the reconstruction loss;
\textit{v}) \textit{Spatio-temporal graph auto-encoder} (STGAE)~\cite{wiederer2022anomaly} that uses convolutional networks temporally and graph convolutional networks spatially, and uses bi-variate Gaussian reconstruction error to build autoencoder to derive anomaly score\footnote{We use the STGAE-biv variant from the paper, since the version with KDE is computationally too expensive. With $n$ trajectories in the training set, and $m$ trajectories in the testing set, the KDE complexity is $O(mn)$. The KDE did not complete given two days of computation time.}; \textit{vi}) DSAB-biv, which is the variant of our model, that uses the same RGAT structure, but uses the bi-variate Gaussian loss as in \cite{wiederer2022anomaly,mohamed2020social}.
Each of the baseline methods produces an anomaly score for the vehicles in each time step within a time window, and the same process is used following Eq~\eqref{eq:car_score} and \eqref{eq:stretch} to calculate the vehicle and scene anomaly scores.

Out of all the baselines, Seq2Seq, RTR, CVM and LTI consider only each vehicle's own trajectory, and STGAE considers the relationships between vehicles. The heuristic models CVM and LTI are able to detect non-free-flow scenarios where the vehicle speed changes dramatically. The tensor PCA can capture linear correlations among trajectories, and the neural-network methods Seq2seq and STGAE can further capture non-linear patterns.

\subsubsection{Metrics}
We follow the practice used in the anomaly detection works~\cite{liu2022benchmarking,ding2021inductive,pei2022resgcn}, and include the following standard metrics: \textit{i}) ROC-AUC score, which is widely used for anomaly detection. \textit{ii}) Average precision, which summarizes the precision-recall curve into a single value. \textit{iii})  Precision@k. In settings where one is only able to e.g., manually verify a fixed number of anomalies, Precision@k measures the relevance of the samples we check.

\subsection{Implementation details} \label{sec:details}

The model setting for DSAB is as follows.  For simulated data: For graph construction, the distance threshold $\delta_x$ is 0.1 miles, and the lane threshold $\delta_l = 1$, i.e., vehicles only attend to its own and immediate neighboring lanes within 0.1 miles upstream and downstream. The window size $T$ is 15s, and sampling interval is 1s. The model hidden size $d_h = 5$, the number of attention heads $K = 3$. In training, the model is trained for 500 epochs, with a batch size of 64. The initial learning rate is 0.05, and decreases by half every 50 epochs. Gradient clipping is used to avoid exploding gradients, with max norm of 1. For abnormal scene detection, the stretch length $\delta_s$ is set to 0.15 miles. 
For the highD data: The window size $T$ is 10s. The distance threshold $\delta_x$ is 0.2 mile. The batch size is set to 256. All other configurations are the same as the simulation data above. 

Our model requires a constant number of vehicles in a single time window.  \edit{For vehicles with incomplete trajectories, due to vehicles entering or leaving the observed stretch, two approaches can be used: \textit{i}) discard the vehicles with incomplete trajectories; or \textit{ii}) linearly extrapolate the trajectories assuming constant velocity dynamics, and then mask the extrapolated part when calculating the loss.  The first approach is used for simulation data on long stretches with a small portion of incomplete trajectories. the second approach  is used for real-world highD data with short stretches, and a large fraction of incomplete trajectories.}




\begin{table}
\centering
\caption{Abnormal \textit{vehicle} detection performance on the test set of comprehensive scenario, where all abnormal behaviors exist on the highway. Our model is the only one that can identify anomalous cars in traffic. }
\label{table:car}
\begin{tabular}{lccccc} 
\toprule
          & Pre@100       & Pre@200        & Pre@500        & Avg Pre        & AUC         \\ 
\midrule
LTI       & 0.21          & 0.205          & 0.210          & 0.093          & 0.623           \\
CVM       & 0.15          & 0.140          & 0.142          & 0.089          & 0.617           \\
RTR       & 0.16          & 0.190          & 0.228          & 0.109          & 0.598           \\
Seq2seq   & 0.56          & 0.435          & 0.310          & 0.132          & 0.770           \\
STGAE     & 0.18          & 0.180          & 0.178          & 0.093          & 0.641           \\
DSAB-biv & 0.15          & 0.150          & 0.114          & 0.062          & 0.655           \\
DSAB (Ours)      & \textbf{0.82} & \textbf{0.775} & \textbf{0.726} & \textbf{0.381} & \textbf{0.900}  \\
\bottomrule
\end{tabular}
\end{table}

\begin{table}
\centering
\caption{Abnormal \textit{scene} detection performance. While identifying abnormal scene is an easier task than  identifying specific abnormal cars, and most method can perform well, ours is still the best in performance, with an increase of 0.1 in ROC-AUC score and average precision.}
\label{table:scene}
\begin{tabular}{lccccc} 
\toprule
          & Pre@100       & Pre@200        & Pre@500        & Avg Pre        & AUC         \\ 
\midrule
LTI       & 0.89          & 0.860          & 0.834          & 0.690          & 0.726           \\
CVM       & 0.67          & 0.625          & 0.640          & 0.662          & 0.714           \\
RTR       & 0.89          & 0.895          & 0.864          & 0.769          & 0.744           \\
Seq2seq   & 0.87          & 0.870          & 0.860          & 0.742          & 0.745           \\
STGAE     & 0.82          & 0.740          & 0.776          & 0.666          & 0.678           \\
DSAB-biv & 0.73          & 0.665          & 0.610          & 0.470          & 0.551           \\
DSAB (Ours)     & \textbf{0.93} & \textbf{0.950} & \textbf{0.912} & \textbf{0.859} & \textbf{0.841}  \\
\bottomrule
\end{tabular}
\end{table}

\subsection{Results and analysis}

In this section, we quantitatively compare our method with the baselines on simulation data. We also conduct ablation studies to see the influence of each model component on performance.
\begin{table*}[]
\centering
\caption{Abnormal car detection performance on each individual anomaly scenario. Our method is the only one that can detect slow and stalled car in traffic, and is also the best at detecting speeding cars. }
\label{table:scenarios}
 \begin{tabular}{l|ccc|ccc|ccc|ccc} 
\toprule
          & \multicolumn{3}{c|}{slow}                       & \multicolumn{3}{c|}{speeding}                   & \multicolumn{3}{c|}{tailgating}                 & \multicolumn{3}{c}{stalled}                      \\ 
\midrule
          & Pre@100       & Avg Pre        & AUC            & Pre@100       & Avg Pre        & AUC            & Pre@100       & Avg Pre        & AUC            & Pre@100       & Avg Pre        & AUC             \\ 
\midrule
LTI       & 0.00          & 0.023          & 0.443          & 0.82          & 0.512          & 0.877          & 0.02          & 0.161          & 0.835          & 0.00          & 0.030          & 0.010           \\
CVM       & 0.01          & 0.023          & 0.432          & 0.67          & 0.472          & 0.871          & 0.06          & \textbf{0.177} & \textbf{0.840} & 0.00          & 0.030          & 0.010           \\
RTR       & 0.00          & 0.030          & 0.466          & 0.61          & 0.365          & 0.655          & 0.08          & 0.102          & 0.400          & 0.00          & 0.046          & 0.410           \\
Seq2seq   & 0.02          & 0.046          & 0.668          & 0.85          & 0.549          & 0.923          & 0.18          & 0.091          & 0.709          & 0.03          & 0.056          & 0.676           \\
STGAE     & 0.01          & 0.048          & 0.573          & 0.83          & 0.335          & 0.712          & 0.01          & 0.055          & 0.491          & 0.00          & 0.025          & 0.454           \\
DSAB-biv & 0.00          & 0.025          & 0.465          & 0.29          & 0.092          & 0.600          & 0.05          & 0.043          & 0.436          & 0.13          & 0.066          & 0.696           \\
DSAB (Ours)      & \textbf{0.97} & \textbf{0.246} & \textbf{0.841} & \textbf{1.00} & \textbf{0.907} & \textbf{0.995} & \textbf{0.33} & 0.148          & 0.747          & \textbf{0.98} & \textbf{0.148} & \textbf{0.762}  \\
\bottomrule
\end{tabular}
\end{table*}

\begin{figure*}[ht]
\centering
\begin{subfigure}{0.19\textwidth}
    \includegraphics[width=\linewidth]{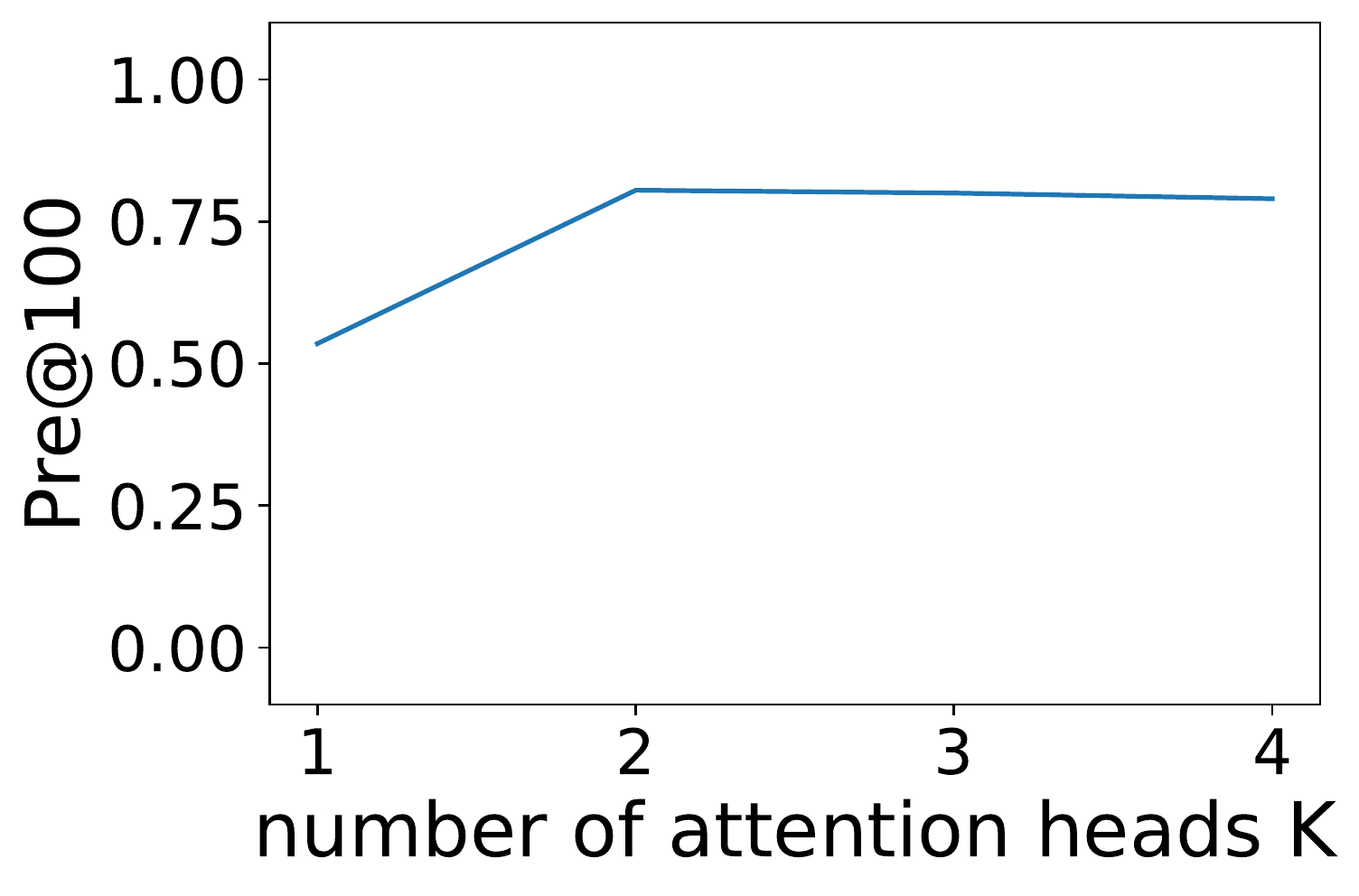}
    \caption{\normalfont{attention heads}}
    \Description{A line chart of Pre\@100 vs the number of attention heads K, where K has four discrete values of 1,2,3 and 4, and Pre\@100 mostly centers at 0.75 but gets a value of 0.5 when there is only one attention head}
    \label{fig:attention_heads}
\end{subfigure}
\begin{subfigure}{0.19\textwidth}
    \includegraphics[width=\linewidth]{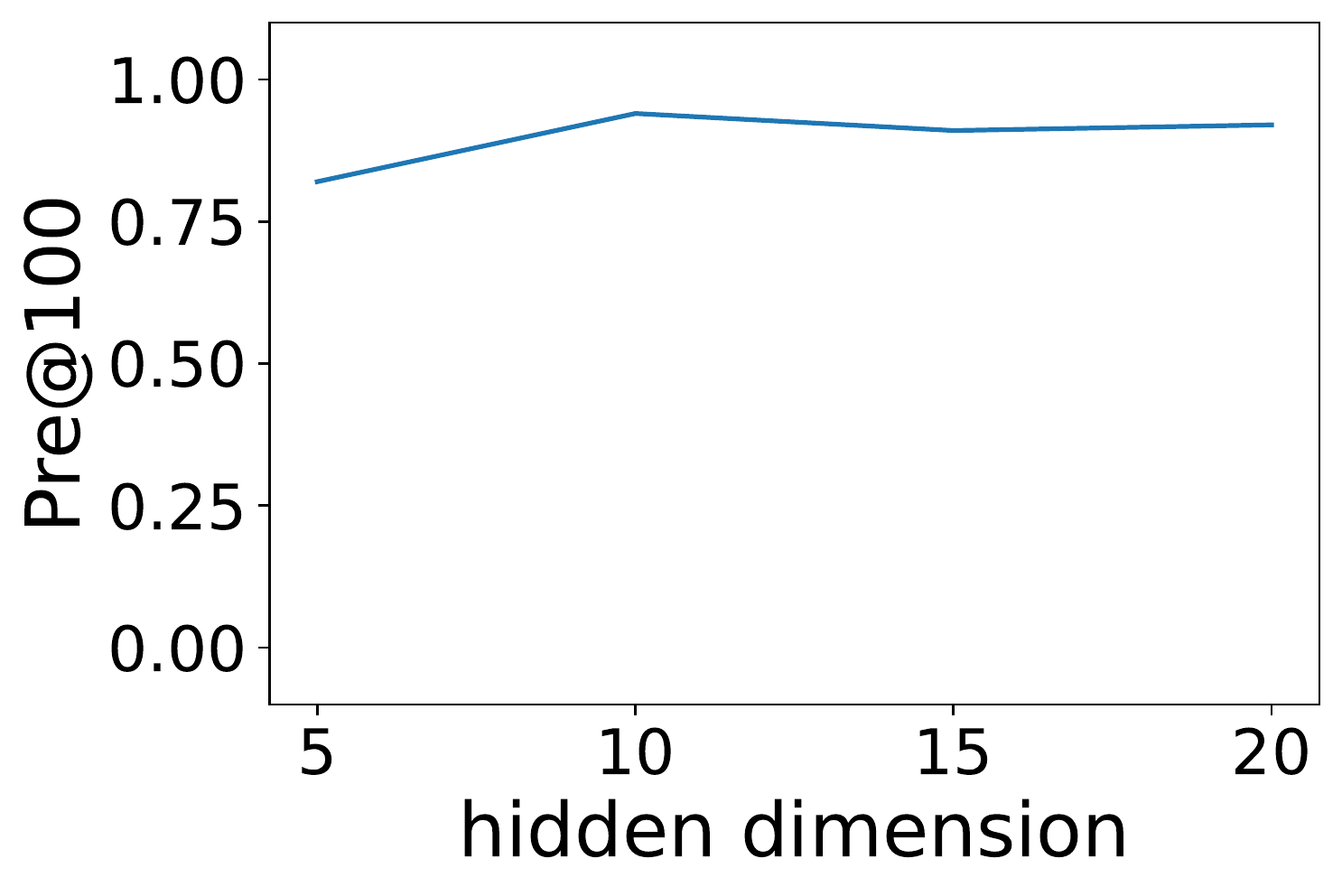}
    \caption{\normalfont{hidden dimension}}
    \Description{A line chart of Pre\@100 vs hidden dimension, where hidden dimension has four discrete values of 5,10,15,20, and Pre\@100 has corresponding values between 0.75 and 1.00. }
    \label{fig:hiddim}
\end{subfigure}
\begin{subfigure}{0.19\textwidth}
    \includegraphics[width=\linewidth]{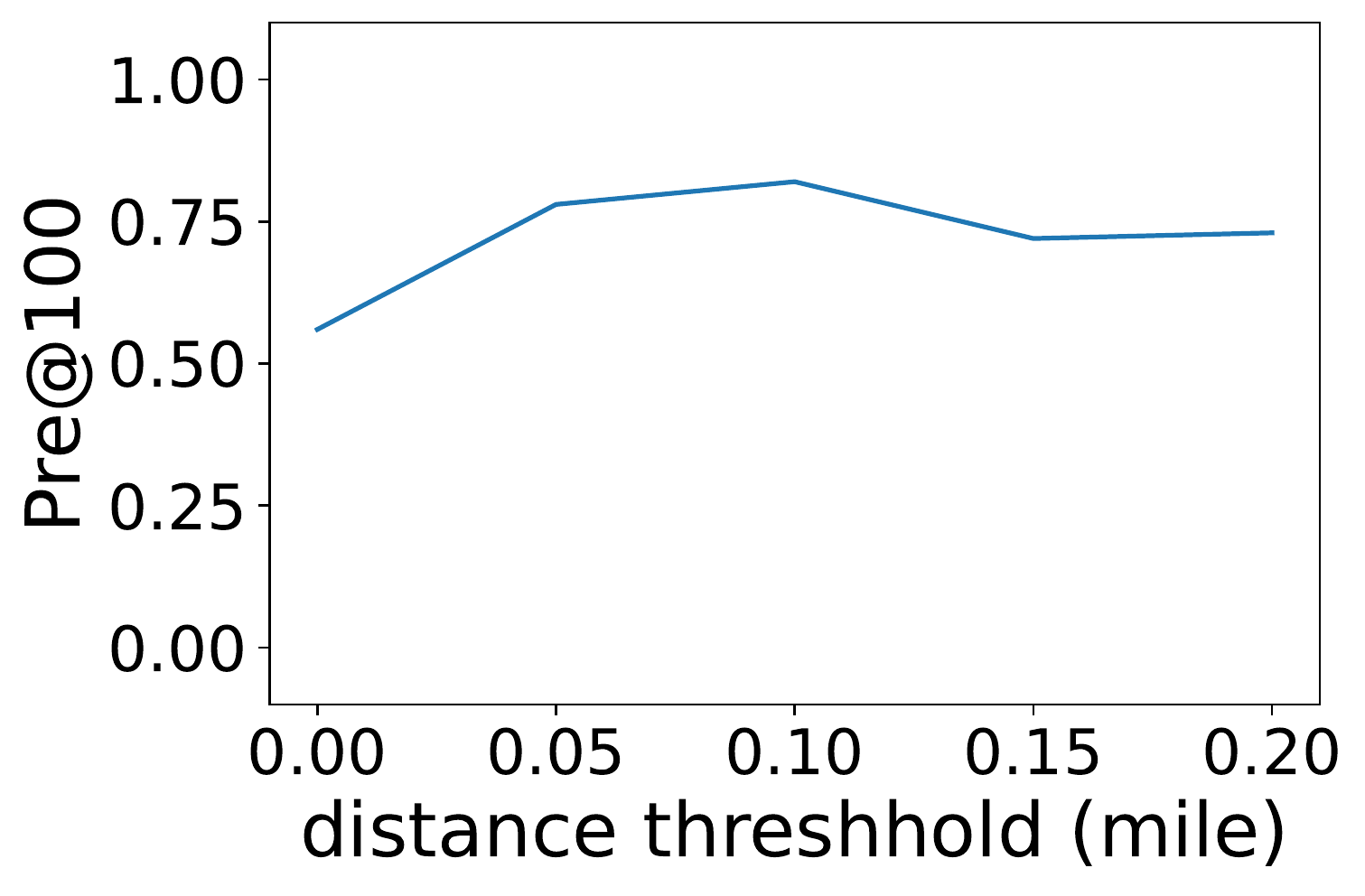}
    \caption{\normalfont{attention distance}}
    \Description{A line chart of Pre\@100 vs attention distance, where attention distance has five discrete values of 0.00,0.05,0.10,0.15,0.20 and Pre\@100 has corresponding values between 0.5 and 0.75.}
    \label{fig:attention_distance_threshhold}
\end{subfigure}
\begin{subfigure}{0.19\textwidth}
    \includegraphics[width=\linewidth]{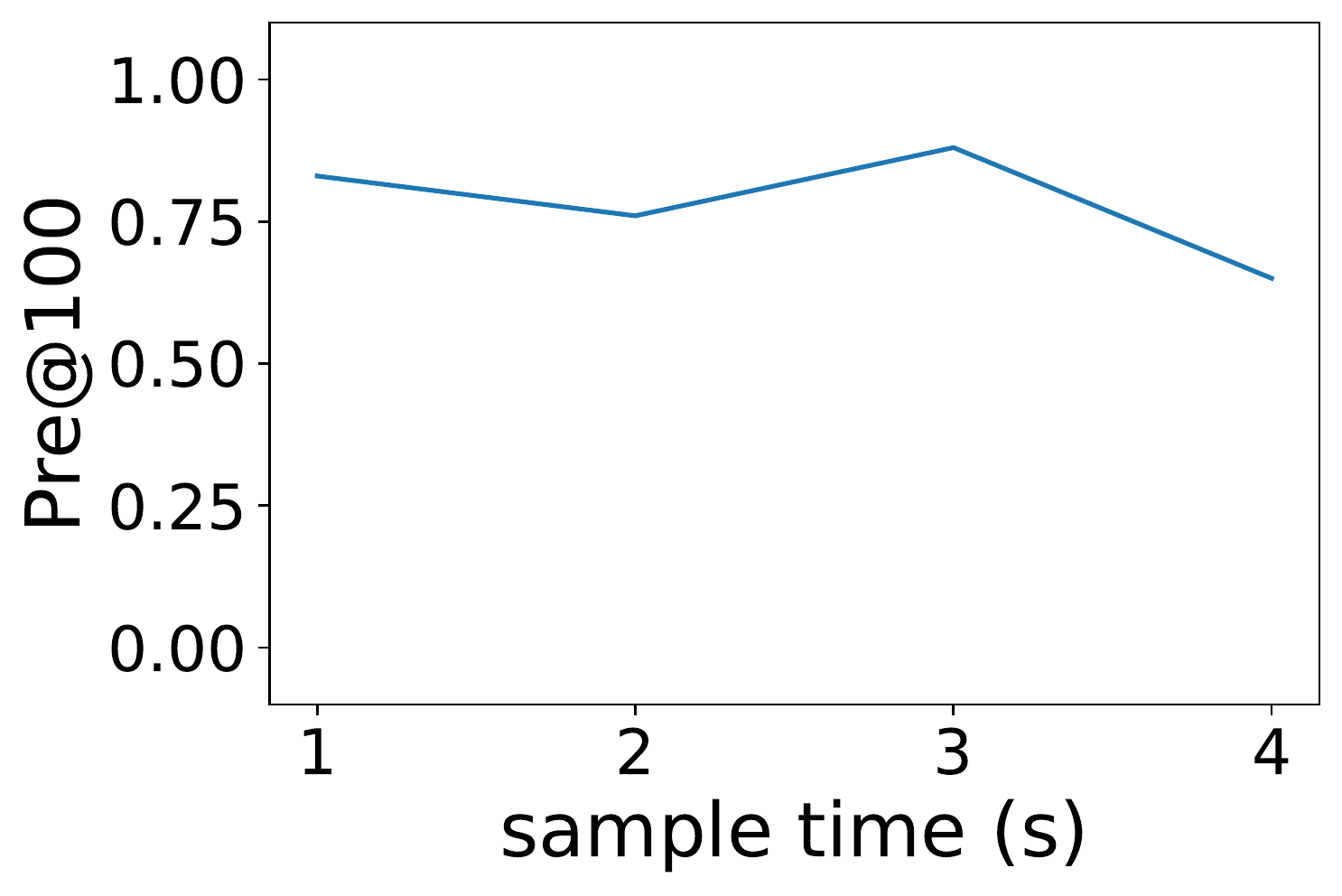}
    \caption{\normalfont{sample time}}
    \Description{A line chart of Pre\@100 vs sample time, where sample time has four discrete values of 1,2,3,4, and Pre\@100 has corresponding values between 0.5 and 1.00.}
    \label{fig:sample_time}
\end{subfigure}
\begin{subfigure}{0.19\textwidth}
    \includegraphics[width=\linewidth]{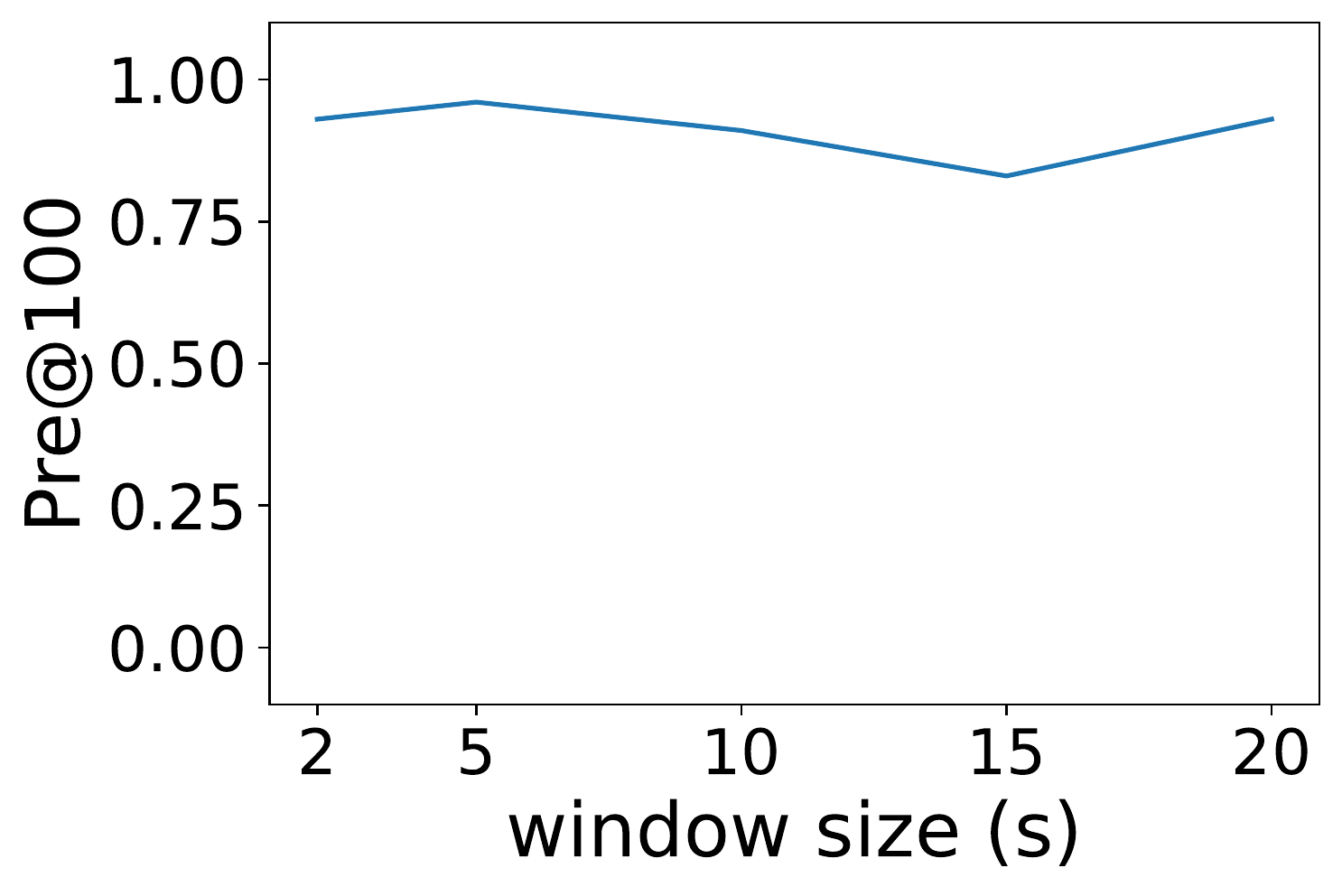}
    \caption{\normalfont{window size}}
    \Description{A line chart of Pre\@100 vs window size, where window size is from 2s to 20s and Pre\@100 has corresponding values between 0.75 and 1.00.}
    \label{fig:window_size}
\end{subfigure}

\caption{Influence of different parameters. The model benefits from having more than one attention head, and is not sensitive to the number of heads and hidden dimensions otherwise. On spatio-temporal graph construction, when attention distance threshhold equals zero, a car does not attend to any neighbor, and the performance significantly drops. Otherwise, the model is not sensitive to the window size and sample frequency, allowing us flexibility to choose parameters.}
\label{fig:ablation_graph}
\end{figure*}

\subsubsection{Model comparison}
We first benchmark the methods on the test set of the comprehensive scenario, where all abnormal behaviors exist on the highway. The result for abnormal \textit{vehicle} detection is shown in Table~\ref{table:car}. We can see that our model is the only one that can identify the specific car that behaves abnormally in the traffic. The precision scores at different k vales are constantly above 0.7 in our method. For the other methods, the precision score are around 0.2 most of the time, and are never higher than 0.6. Our average precision is also around 0.3 higher than the second best method. The ROC-AUC score for our method is 0.9, which is more than 0.1 higher than the second best method. This shows that without consideration of neighboring vehicles,
Seq2Seq, RTR, CVM and LTI are unable to detect the social anomalies. Meanwhile, although STGAE and DSAB-biv both model the neighboring vehicle interactions via graph neural networks, they use bi-viariate Gaussian trajectory loss and cannot well capture the discrete lane-changing behaviors, and have similarly poor results. 

We also report the performance of abnormal \textit{scene} detection for all method, shown in Table~\ref{table:scene}. We can see that the performance of all methods have a significant improvement. This is because  detecting abnormal scenes is easier task than detecting the specific abnormal vehicles. For example, a slow or stalled car can cause its following cars to brake abruptly, which baseline models like CVM are able to detect. Moreover, when aggregating a group of vehicles over space and time, the abnormal rate is higher, and the methods work much better on a balanced dataset. Nonetheless, our method is still the best,  with an increase of 0.1 in AUC score and average precision.

\subsubsection{Performance on individual anomaly scenarios}
Next, we investigate the model performance on each individual anomaly scenario. The result is shown in Table~\ref{table:scenarios}. We can see that our method is the only one that can detect slow and stalled cars in traffic, with Precision@100 score above 0.9, compared with scores constantly less than 0.1 for other methods. While all methods perform relatively well at detecting speeding cars,  our method is the best, with an improvement of 0.1 in Precision@100, and a ROC-AUC score near 1. Our method is relatively less effective in detecting tailgating cars, yet the Precision@100 is still the highest among all methods, and no method is constantly better than ours in every metric. We note that baseline methods can detect tailgating cars, because they have different car-following dynamics as described in section~\ref{sec:dataset}.

\subsubsection{Ablation study}
In this section, we examine the influence of different components of the model, as well as model sensitivity to the configurations of the spatio-temporal graph data.

First, we study the influence of the model parameters, with results shown in Fig~\ref{fig:attention_heads} and \ref{fig:hiddim}. The influence of number of attention heads is shown in Fig~\ref{fig:attention_heads}. We can see that the model benefits from having more than one attention head, while being not sensitive when the number of heads is larger than one. Meanwhile, Fig~\ref{fig:hiddim} indicates that the model is not sensitive to the hidden dimension. Thus we choose 5 as hidden dimension for a smaller model size.

Next, we study the influence of spatio-temporal graph parameters, with results shown in Fig~\ref{fig:attention_distance_threshhold},  \ref{fig:window_size} and \ref{fig:sample_time}. Spatially, the threshold of attention distance determines how far away in distance one vehicle attends to as its neighboring car. A threshold of zero means a car do not attend to any neighbor, and consequently the spatial graph is not used. We can see in Fig~\ref{fig:attention_distance_threshhold} that when the threshold equals zero, the performance significantly drops by 0.3 in Precision@100. When  threshold is larger than 0.05 miles, the performance is not sensitive to the threshold value. The result indicates that the neighboring vehicles within 0.05 miles around the ego vehicle is the most important. Temporally, we study the influence of the time window size and the sample frequency. We first fix the sampling interval at 1s and vary the window size from 2s to 20s, then fix the window size at 15s and vary the sampling interval from 1s to 4s. Fig~\ref{fig:window_size} and Fig~\ref{fig:sample_time} show the results. We can see that the model performs best when interval is less or equal to 3s, and is not sensitive to the window size and sample frequency otherwise. This allows us flexibility to choose a larger window size and coarser sampling frequency to aid computational efficiency.


\subsection{Qualitative Results}
In this section, we qualitatively show the performance of our model on real-world HighD data. The traffic in this data has unique characteristics, and our model is able to learn the norms and capture the anomalies that deviates from the norm. 

\begin{figure}
\centering
\begin{subfigure}[t]{\columnwidth}
    \includegraphics[width=\linewidth]{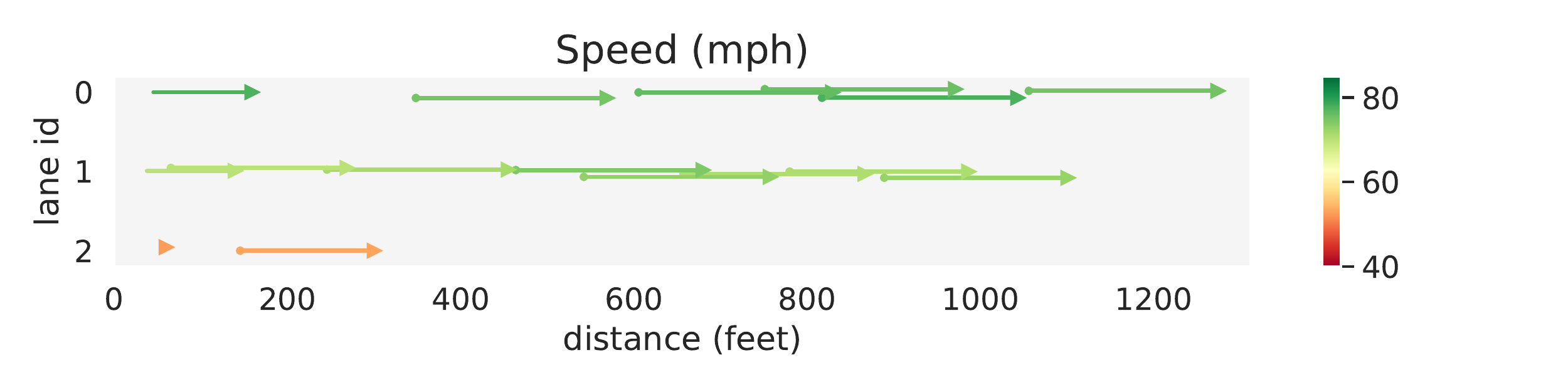}
    \caption{\normalfont{ Normal traffic condition. The typical speed varies by lane. From leftmost to rightmost lane, the average speed are 76, 69 and 56 mph respectively, calculated from training data}}
    \Description{A rectangular figure containing vehicle trajectories with x-axis as distance and y-axis as lane id. There are multiple lines on the figure, each of which denotes the trajectory of a single vehicle, with a dot denoting the starting point
    and triangle denoting the endpoint. The lines has different colors to represent the vehicle speed, where the color bar is roughly from red to yellow to green and represents 40 to 60 to 80. There are several green lines on lane 0, several light green lines on lane1 and two orange lines on lane 2.}
    \label{fig:highD_norm}
\end{subfigure}

\begin{subfigure}[t]{\columnwidth}
    \includegraphics[width=\linewidth]{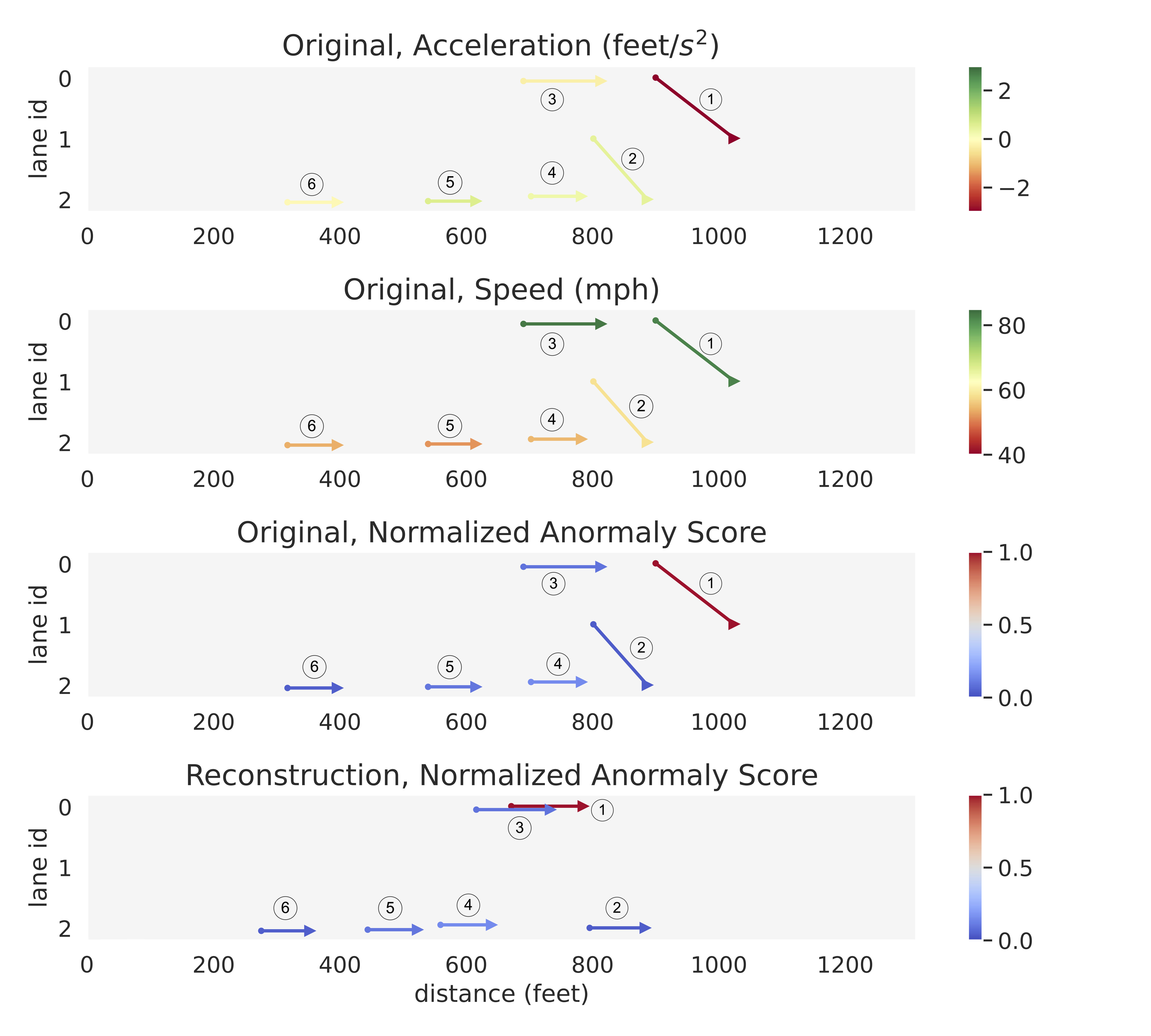}
     \caption{\normalfont{Vehicle with largest anomaly score. The abnormal vehicle 1 is cutting in front of vehicle 2, while having dramatic deceleration, forcing vehicle 2 to change lane as well.}}
    \Description{Four rectangular figures containing vehicle trajectories with x-axis as distance and y-axis as lane id. There are multiple lines on the figure, each of which denotes the trajectory of a single vehicle, with a dot denoting the starting point and a triangle denoting the endpoint. The lines has different colors to represent different features. For the first top figure, the color represents vehicle acceleration where the color bar is from red to yellow to green and represents -2 to 0 to 2. There is one red line with starting point at lane 0 but end point at lane 1. The rest lines have relatively more yellow or more green colors. For the second top figure, the color represents vehicle speed where the color bar is from red to yellow to green and represents 40 to 60 to 80. All lines have a similar pattern with the first top figure but with different colors. For the third figure, the color represents the original normalized anomaly score of each vehicle where the color bar is from blue to white to red and represents 0 to 0.5 to 1. The red line mentioned in the first top figure also has red color here while the rest lines are all blue. For the fourth figure, the color represents the reconstructed normalized anomaly score of each vehicle where the color bar is from blue to white to red and represents 0 to 0.5 to 1. The red line mentioned in the first top figure also has red color here while the rest lines are all blue.}
    \label{fig:highD_abn1}
\end{subfigure}

\quad
\begin{subfigure}[t]{\columnwidth}
    \includegraphics[width=\linewidth]{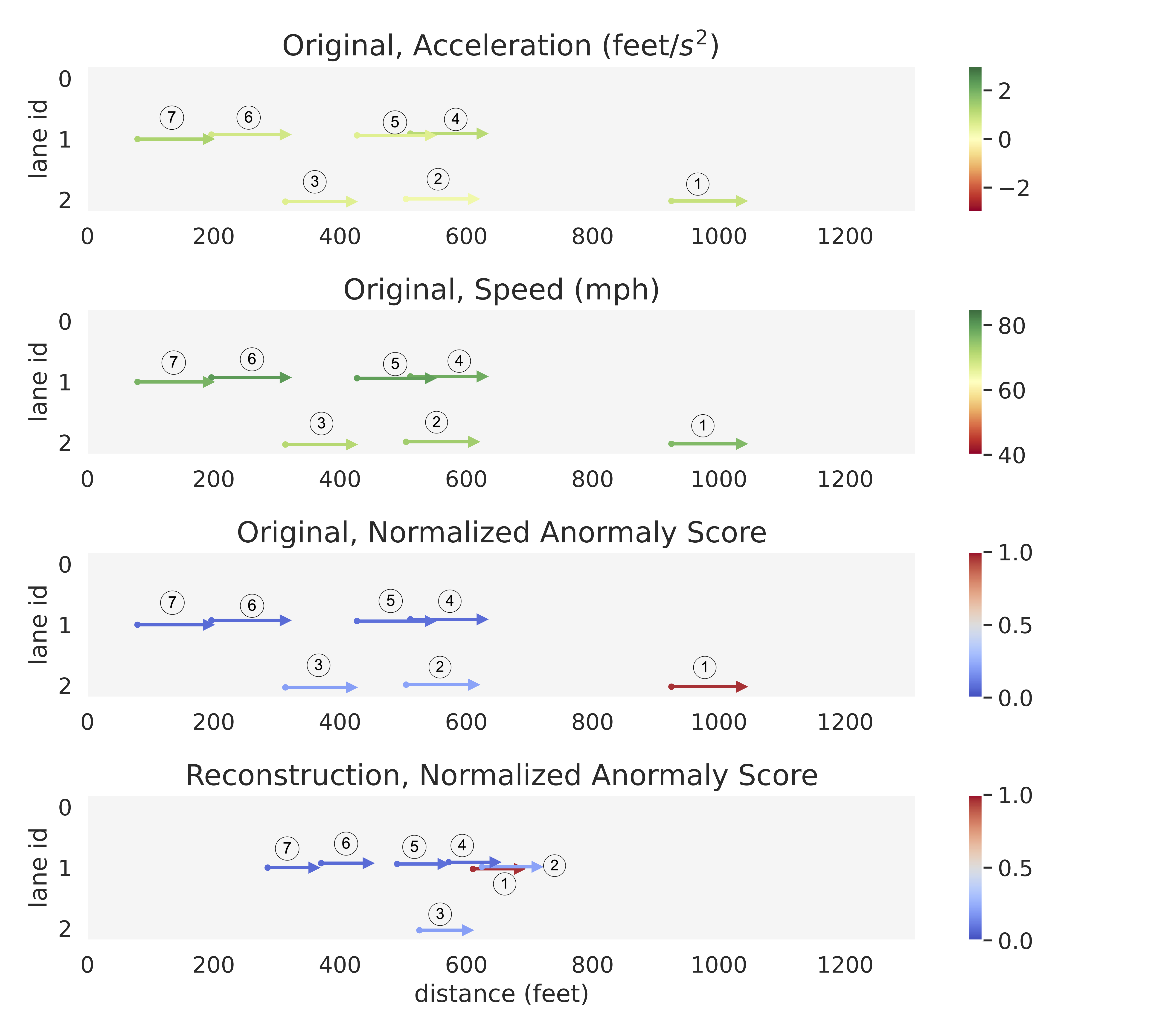}
    \caption{\normalfont{Vehicle with second largest anomaly score. The abnormal vehicle 1 is driving too fast with respect to the lane it is in. The reconstruction actually puts vehicle 1 in the middle lane which has larger typical speed.} }
    \Description{Four rectangular figures containing vehicle trajectories with x-axis as distance and y-axis as lane id. There are multiple lines on the figure, each of which denotes the trajectory of a single vehicle, with a dot denoting the starting point
    and triangle denoting the endpoint. The lines has different colors to represent different features. For the first top figure, the color represents vehicle acceleration where the color bar is from red to yellow to green and represents -2 to 0 to 2. There is one green line with index of vehicle 1 on lane 2. The rest lines also have green colors. For the second top figure, the color represents vehicle speed where the color bar is from red to yellow to green and represents 40 to 60 to 80. There is one green line with index of vehicle 1 on lane 2. The rest lines also have green colors. For the third figure, the color represents the original normalized anomaly score of each vehicle where the color bar is from blue to white to red and represents 0 to 0.5 to 1. The line with index of vehicle 1 in the first top figure has red color here while the rest lines are all blue. For the fourth figure, the color represents the reconstructed normalized anomaly score of each vehicle where the color bar is from blue to white to red and represents 0 to 0.5 to 1. The line with index of vehicle 1 in the first top figure has red color here while the rest lines are all blue.}
    \label{fig:highD_abn2}
\end{subfigure}

\caption{Qualitative study of real-world HighD traffic data}
\label{fig:highD}
\end{figure}

Specifically, the speed limit for German highways is very loose, at least 75 mph, or no speed limits in some parts. And it is not uncommon to observe speeds larger than 80 mph. Thus, speeding is not ranked among the largest anomalies. On the other hand, the typical traffic speed varies significantly by lane, as shown in Fig~\ref{fig:highD_norm}. From leftmost to rightmost lane, the average speed are 76, 69 and 56 mph respectively, calculated from training data. That is to say, the faster the car, the more to the left the car tends to be. Accordingly, the vehicle with much higher speed than its corresponding lane is detected as abnormal, even though the speed is absolutely normal when lanes are not considered. We manually inspect the vehicles with top anomaly scores, and found them to be abnormal either because of aggressive driving with dramatic acceleration/deceleration, or because of speed range violation with respect to the lane.

We show some of the top anomalies in Fig~\ref{fig:highD}. Each line denotes the trajectory of a single car, with a dot denoting the starting point and triangle the end point. We add small perturbations laterally to aid visualization. The speed, acceleration and vehicle anomaly score at the corresponding time are shown. The reconstruction  shows the trajectories with the largest probability. Fig~\ref{fig:highD_abn1} shows the vehicle with largest anomaly score. The abnormal vehicle 1 is cutting in front of vehicle 2, while having dramatic deceleration, forcing the other car to change lane as well. Fig~\ref{fig:highD_abn2} shows the vehicle with second largest anomaly score. The abnormal vehicle 1 is driving too fast with respect to the lane it is in -- 20 mph larger than the typical speed in rightmost lane. The reconstruction actually puts vehicle 1 in the middle lane which has larger typical speed.

While we presented the top two anomalies as identified by DSAB, with more examples in  Appendix~\ref{sec:highD_more}, the result illustrates that the method is applicable to real world datasets.

 

\section{Conclusion} ~\label{Sec:conclude}
Advanced connectivity and sensing will continue to transform intelligent traffic management. In this work, we tackle an important problem of abnormal driving behavior detection, using trajectories produced by IoT highway video surveillance systems. Specifically we detect the exact anomalous vehicles considering the vehicular interaction with DSAB, an autoencoder based on Recurrent Graph Attention Networks. The results demonstrate the method captures the spatial-temporal trajectory dynamics, while considering both the neighbor interactions and the stocasticity in driving behaviors. Extensive experiments on both simulation and
real-world data sets show the ability of our model to
scale to large highway monitoring systems with thousands
of vehicles, and detect a variety of abnormal behaviors. The performance on identifying single vehicle anomalies is state of the art, indicating potential to pinpoint specific problematic vehicles in a large traffic stream. 

\noindent\textbf{Acknowledgments}
This work is supported by the National Science Foundation (NSF) under Grant No. CIS-2033580 and the USDOT Dwight D. Eisenhower Fellowship program under Grant No. 693JJ322NF5201.

\bibliographystyle{ACM-Reference-Format}
\bibliography{ref}


\begin{thebibliography}{55}


\ifx \showCODEN    \undefined \def \showCODEN     #1{\unskip}     \fi
\ifx \showDOI      \undefined \def \showDOI       #1{#1}\fi
\ifx \showISBNx    \undefined \def \showISBNx     #1{\unskip}     \fi
\ifx \showISBNxiii \undefined \def \showISBNxiii  #1{\unskip}     \fi
\ifx \showISSN     \undefined \def \showISSN      #1{\unskip}     \fi
\ifx \showLCCN     \undefined \def \showLCCN      #1{\unskip}     \fi
\ifx \shownote     \undefined \def \shownote      #1{#1}          \fi
\ifx \showarticletitle \undefined \def \showarticletitle #1{#1}   \fi
\ifx \showURL      \undefined \def \showURL       {\relax}        \fi
\providecommand\bibfield[2]{#2}
\providecommand\bibinfo[2]{#2}
\providecommand\natexlab[1]{#1}
\providecommand\showeprint[2][]{arXiv:#2}

\bibitem[Ahmed(1999)]%
        {ahmed1999modeling}
\bibfield{author}{\bibinfo{person}{Kazi~Iftekhar Ahmed}.}
  \bibinfo{year}{1999}\natexlab{}.
\newblock \emph{\bibinfo{title}{Modeling drivers' acceleration and lane
  changing behavior}}.
\newblock \bibinfo{thesistype}{Ph.\,D. Dissertation}.
  \bibinfo{school}{Massachusetts Institute of Technology}.
\newblock


\bibitem[Alahi et~al\mbox{.}(2016)]%
        {alahi2016social}
\bibfield{author}{\bibinfo{person}{Alexandre Alahi}, \bibinfo{person}{Kratarth
  Goel}, \bibinfo{person}{Vignesh Ramanathan}, \bibinfo{person}{Alexandre
  Robicquet}, \bibinfo{person}{Li Fei-Fei}, {and} \bibinfo{person}{Silvio
  Savarese}.} \bibinfo{year}{2016}\natexlab{}.
\newblock \showarticletitle{Social lstm: Human trajectory prediction in crowded
  spaces}. In \bibinfo{booktitle}{\emph{Proceedings of the IEEE conference on
  computer vision and pattern recognition}}. \bibinfo{pages}{961--971}.
\newblock


\bibitem[Alkinani et~al\mbox{.}(2020)]%
        {alkinani2020detecting}
\bibfield{author}{\bibinfo{person}{Monagi~H Alkinani},
  \bibinfo{person}{Wazir~Zada Khan}, {and} \bibinfo{person}{Quratulain
  Arshad}.} \bibinfo{year}{2020}\natexlab{}.
\newblock \showarticletitle{Detecting human driver inattentive and aggressive
  driving behavior using deep learning: Recent advances, requirements and open
  challenges}.
\newblock \bibinfo{journal}{\emph{Ieee Access}}  \bibinfo{volume}{8}
  (\bibinfo{year}{2020}), \bibinfo{pages}{105008--105030}.
\newblock


\bibitem[Bai et~al\mbox{.}(2019)]%
        {bai2019traffic}
\bibfield{author}{\bibinfo{person}{Shuai Bai}, \bibinfo{person}{Zhiqun He},
  \bibinfo{person}{Yu Lei}, \bibinfo{person}{Wei Wu}, \bibinfo{person}{Chengkai
  Zhu}, \bibinfo{person}{Ming Sun}, {and} \bibinfo{person}{Junjie Yan}.}
  \bibinfo{year}{2019}\natexlab{}.
\newblock \showarticletitle{Traffic anomaly detection via perspective map based
  on spatial-temporal information matrix.}. In \bibinfo{booktitle}{\emph{CVPR
  Workshops}}. \bibinfo{pages}{117--124}.
\newblock


\bibitem[Bandyopadhyay et~al\mbox{.}(2020)]%
        {bandyopadhyay2020outlier}
\bibfield{author}{\bibinfo{person}{Sambaran Bandyopadhyay},
  \bibinfo{person}{Saley~Vishal Vivek}, {and} \bibinfo{person}{MN Murty}.}
  \bibinfo{year}{2020}\natexlab{}.
\newblock \showarticletitle{Outlier resistant unsupervised deep architectures
  for attributed network embedding}. In \bibinfo{booktitle}{\emph{Proceedings
  of the 13th international conference on web search and data mining}}.
  \bibinfo{pages}{25--33}.
\newblock


\bibitem[Brody et~al\mbox{.}(2021)]%
        {brody2021attentive}
\bibfield{author}{\bibinfo{person}{Shaked Brody}, \bibinfo{person}{Uri Alon},
  {and} \bibinfo{person}{Eran Yahav}.} \bibinfo{year}{2021}\natexlab{}.
\newblock \showarticletitle{How attentive are graph attention networks?}
\newblock \bibinfo{journal}{\emph{arXiv preprint arXiv:2105.14491}}
  (\bibinfo{year}{2021}).
\newblock


\bibitem[Cai et~al\mbox{.}(2021)]%
        {cai2021structural}
\bibfield{author}{\bibinfo{person}{Lei Cai}, \bibinfo{person}{Zhengzhang Chen},
  \bibinfo{person}{Chen Luo}, \bibinfo{person}{Jiaping Gui},
  \bibinfo{person}{Jingchao Ni}, \bibinfo{person}{Ding Li}, {and}
  \bibinfo{person}{Haifeng Chen}.} \bibinfo{year}{2021}\natexlab{}.
\newblock \showarticletitle{Structural temporal graph neural networks for
  anomaly detection in dynamic graphs}. In
  \bibinfo{booktitle}{\emph{Proceedings of the 30th ACM international
  conference on Information \& Knowledge Management}}.
  \bibinfo{pages}{3747--3756}.
\newblock


\bibitem[Carlos et~al\mbox{.}(2018)]%
        {carlos2018evaluation}
\bibfield{author}{\bibinfo{person}{Manuel~Ricardo Carlos},
  \bibinfo{person}{Mario~Ezra Arag{\'o}n}, \bibinfo{person}{Luis~C
  Gonz{\'a}lez}, \bibinfo{person}{Hugo~Jair Escalante}, {and}
  \bibinfo{person}{Fernando Mart{\'\i}nez}.} \bibinfo{year}{2018}\natexlab{}.
\newblock \showarticletitle{Evaluation of detection approaches for road
  anomalies based on accelerometer readings—Addressing who’s who}.
\newblock \bibinfo{journal}{\emph{IEEE Transactions on Intelligent
  Transportation Systems}} \bibinfo{volume}{19}, \bibinfo{number}{10}
  (\bibinfo{year}{2018}), \bibinfo{pages}{3334--3343}.
\newblock


\bibitem[Chen et~al\mbox{.}(2021)]%
        {chen2021dual}
\bibfield{author}{\bibinfo{person}{Jingyuan Chen}, \bibinfo{person}{Guanchen
  Ding}, \bibinfo{person}{Yuchen Yang}, \bibinfo{person}{Wenwei Han},
  \bibinfo{person}{Kangmin Xu}, \bibinfo{person}{Tianyi Gao},
  \bibinfo{person}{Zhe Zhang}, \bibinfo{person}{Wanping Ouyang},
  \bibinfo{person}{Hao Cai}, {and} \bibinfo{person}{Zhenzhong Chen}.}
  \bibinfo{year}{2021}\natexlab{}.
\newblock \showarticletitle{Dual-modality vehicle anomaly detection via
  bilateral trajectory tracing}. In \bibinfo{booktitle}{\emph{Proceedings of
  the IEEE/CVF Conference on Computer Vision and Pattern Recognition}}.
  \bibinfo{pages}{4016--4025}.
\newblock


\bibitem[Cho et~al\mbox{.}(2014)]%
        {cho2014properties}
\bibfield{author}{\bibinfo{person}{Kyunghyun Cho}, \bibinfo{person}{Bart van
  Merri{\"e}nboer}, \bibinfo{person}{Dzmitry Bahdanau}, {and}
  \bibinfo{person}{Yoshua Bengio}.} \bibinfo{year}{2014}\natexlab{}.
\newblock \showarticletitle{On the Properties of Neural Machine Translation:
  Encoder--Decoder Approaches}.
\newblock \bibinfo{journal}{\emph{Syntax, Semantics and Structure in
  Statistical Translation}} (\bibinfo{year}{2014}), \bibinfo{pages}{103}.
\newblock


\bibitem[Chung et~al\mbox{.}(2014)]%
        {chung2014empirical}
\bibfield{author}{\bibinfo{person}{Junyoung Chung}, \bibinfo{person}{Caglar
  Gulcehre}, \bibinfo{person}{Kyunghyun Cho}, {and} \bibinfo{person}{Yoshua
  Bengio}.} \bibinfo{year}{2014}\natexlab{}.
\newblock \showarticletitle{Empirical evaluation of gated recurrent neural
  networks on sequence modeling}. In \bibinfo{booktitle}{\emph{NIPS 2014
  Workshop on Deep Learning, December 2014}}.
\newblock


\bibitem[Defferrard et~al\mbox{.}(2016)]%
        {defferrard2016convolutional}
\bibfield{author}{\bibinfo{person}{Micha{\"e}l Defferrard},
  \bibinfo{person}{Xavier Bresson}, {and} \bibinfo{person}{Pierre
  Vandergheynst}.} \bibinfo{year}{2016}\natexlab{}.
\newblock \showarticletitle{Convolutional neural networks on graphs with fast
  localized spectral filtering}.
\newblock \bibinfo{journal}{\emph{Advances in neural information processing
  systems}}  \bibinfo{volume}{29} (\bibinfo{year}{2016}).
\newblock


\bibitem[Ding et~al\mbox{.}(2021)]%
        {ding2021inductive}
\bibfield{author}{\bibinfo{person}{Kaize Ding}, \bibinfo{person}{Jundong Li},
  \bibinfo{person}{Nitin Agarwal}, {and} \bibinfo{person}{Huan Liu}.}
  \bibinfo{year}{2021}\natexlab{}.
\newblock \showarticletitle{Inductive anomaly detection on attributed
  networks}. In \bibinfo{booktitle}{\emph{Proceedings of the Twenty-Ninth
  International Conference on International Joint Conferences on Artificial
  Intelligence}}. \bibinfo{pages}{1288--1294}.
\newblock


\bibitem[Ding et~al\mbox{.}(2019)]%
        {ding2019deep}
\bibfield{author}{\bibinfo{person}{Kaize Ding}, \bibinfo{person}{Jundong Li},
  \bibinfo{person}{Rohit Bhanushali}, {and} \bibinfo{person}{Huan Liu}.}
  \bibinfo{year}{2019}\natexlab{}.
\newblock \showarticletitle{Deep anomaly detection on attributed networks}. In
  \bibinfo{booktitle}{\emph{Proceedings of the 2019 SIAM International
  Conference on Data Mining}}. SIAM, \bibinfo{pages}{594--602}.
\newblock


\bibitem[Fazeen et~al\mbox{.}(2012)]%
        {fazeen2012safe}
\bibfield{author}{\bibinfo{person}{Mohamed Fazeen}, \bibinfo{person}{Brandon
  Gozick}, \bibinfo{person}{Ram Dantu}, \bibinfo{person}{Moiz Bhukhiya}, {and}
  \bibinfo{person}{Marta~C Gonz{\'a}lez}.} \bibinfo{year}{2012}\natexlab{}.
\newblock \showarticletitle{Safe driving using mobile phones}.
\newblock \bibinfo{journal}{\emph{IEEE Transactions on Intelligent
  Transportation Systems}} \bibinfo{volume}{13}, \bibinfo{number}{3}
  (\bibinfo{year}{2012}), \bibinfo{pages}{1462--1468}.
\newblock


\bibitem[Gatteschi et~al\mbox{.}(2021)]%
        {gatteschi2021comparing}
\bibfield{author}{\bibinfo{person}{Valentina Gatteschi},
  \bibinfo{person}{Alberto Cannav{\`o}}, \bibinfo{person}{Fabrizio Lamberti},
  \bibinfo{person}{Lia Morra}, {and} \bibinfo{person}{Paolo Montuschi}.}
  \bibinfo{year}{2021}\natexlab{}.
\newblock \showarticletitle{Comparing Algorithms for Aggressive Driving Event
  Detection Based on Vehicle Motion Data}.
\newblock \bibinfo{journal}{\emph{IEEE Transactions on Vehicular Technology}}
  \bibinfo{volume}{71}, \bibinfo{number}{1} (\bibinfo{year}{2021}),
  \bibinfo{pages}{53--68}.
\newblock


\bibitem[Gloudemans et~al\mbox{.}(2023)]%
        {motionSystem}
\bibfield{author}{\bibinfo{person}{Derek Gloudemans}, \bibinfo{person}{Yanbing
  Wang}, \bibinfo{person}{Junyi Ji}, \bibinfo{person}{Gergely Zachar},
  \bibinfo{person}{Will Barbour}, {and} \bibinfo{person}{Daniel~B. Work}.}
  \bibinfo{year}{2023}\natexlab{}.
\newblock \bibinfo{title}{I-24 MOTION: An instrument for freeway traffic
  science}.
\newblock
\newblock
\urldef\tempurl%
\url{https://doi.org/10.48550/ARXIV.2301.11198}
\showDOI{\tempurl}


\bibitem[Gupta et~al\mbox{.}(2018)]%
        {gupta2018social}
\bibfield{author}{\bibinfo{person}{Agrim Gupta}, \bibinfo{person}{Justin
  Johnson}, \bibinfo{person}{Li Fei-Fei}, \bibinfo{person}{Silvio Savarese},
  {and} \bibinfo{person}{Alexandre Alahi}.} \bibinfo{year}{2018}\natexlab{}.
\newblock \showarticletitle{Social gan: Socially acceptable trajectories with
  generative adversarial networks}. In \bibinfo{booktitle}{\emph{Proceedings of
  the IEEE conference on computer vision and pattern recognition}}.
  \bibinfo{pages}{2255--2264}.
\newblock


\bibitem[Hu et~al\mbox{.}(2022)]%
        {hu2022detecting}
\bibfield{author}{\bibinfo{person}{Yue Hu}, \bibinfo{person}{Ao Qu}, {and}
  \bibinfo{person}{Dan Work}.} \bibinfo{year}{2022}\natexlab{}.
\newblock \showarticletitle{Detecting extreme traffic events via a context
  augmented graph autoencoder}.
\newblock \bibinfo{journal}{\emph{ACM Transactions on Intelligent Systems and
  Technology (TIST)}} \bibinfo{volume}{13}, \bibinfo{number}{6}
  (\bibinfo{year}{2022}), \bibinfo{pages}{1--23}.
\newblock


\bibitem[Hu and Work(2020)]%
        {hu2020robust}
\bibfield{author}{\bibinfo{person}{Yue Hu} {and} \bibinfo{person}{Daniel~B
  Work}.} \bibinfo{year}{2020}\natexlab{}.
\newblock \showarticletitle{Robust Tensor Recovery with Fiber Outliers for
  Traffic Events}.
\newblock \bibinfo{journal}{\emph{ACM Transactions on Knowledge Discovery from
  Data (TKDD)}} \bibinfo{volume}{15}, \bibinfo{number}{1}
  (\bibinfo{year}{2020}), \bibinfo{pages}{1--27}.
\newblock


\bibitem[Huang et~al\mbox{.}(2019)]%
        {huang2019stgat}
\bibfield{author}{\bibinfo{person}{Yingfan Huang}, \bibinfo{person}{Huikun Bi},
  \bibinfo{person}{Zhaoxin Li}, \bibinfo{person}{Tianlu Mao}, {and}
  \bibinfo{person}{Zhaoqi Wang}.} \bibinfo{year}{2019}\natexlab{}.
\newblock \showarticletitle{Stgat: Modeling spatial-temporal interactions for
  human trajectory prediction}. In \bibinfo{booktitle}{\emph{Proceedings of the
  IEEE/CVF International Conference on Computer Vision}}.
  \bibinfo{pages}{6272--6281}.
\newblock


\bibitem[Jeni et~al\mbox{.}(2013)]%
        {jeni2013facing}
\bibfield{author}{\bibinfo{person}{L{\'a}szl{\'o}~A Jeni},
  \bibinfo{person}{Jeffrey~F Cohn}, {and} \bibinfo{person}{Fernando
  De~La~Torre}.} \bibinfo{year}{2013}\natexlab{}.
\newblock \showarticletitle{Facing imbalanced data--recommendations for the use
  of performance metrics}. In \bibinfo{booktitle}{\emph{2013 Humaine
  association conference on affective computing and intelligent interaction}}.
  IEEE, \bibinfo{pages}{245--251}.
\newblock


\bibitem[Kipf and Welling(2016)]%
        {kipf2016semi}
\bibfield{author}{\bibinfo{person}{Thomas~N Kipf} {and} \bibinfo{person}{Max
  Welling}.} \bibinfo{year}{2016}\natexlab{}.
\newblock \showarticletitle{Semi-supervised classification with graph
  convolutional networks}.
\newblock \bibinfo{journal}{\emph{arXiv preprint arXiv:1609.02907}}
  (\bibinfo{year}{2016}).
\newblock


\bibitem[Krajewski et~al\mbox{.}(2018)]%
        {highDdataset}
\bibfield{author}{\bibinfo{person}{Robert Krajewski}, \bibinfo{person}{Julian
  Bock}, \bibinfo{person}{Laurent Kloeker}, {and} \bibinfo{person}{Lutz
  Eckstein}.} \bibinfo{year}{2018}\natexlab{}.
\newblock \showarticletitle{The highD Dataset: A Drone Dataset of Naturalistic
  Vehicle Trajectories on German Highways for Validation of Highly Automated
  Driving Systems}. In \bibinfo{booktitle}{\emph{2018 21st International
  Conference on Intelligent Transportation Systems (ITSC)}}.
  \bibinfo{pages}{2118--2125}.
\newblock
\urldef\tempurl%
\url{https://doi.org/10.1109/ITSC.2018.8569552}
\showDOI{\tempurl}


\bibitem[Lee et~al\mbox{.}(2017)]%
        {lee2017desire}
\bibfield{author}{\bibinfo{person}{Namhoon Lee}, \bibinfo{person}{Wongun Choi},
  \bibinfo{person}{Paul Vernaza}, \bibinfo{person}{Christopher~B Choy},
  \bibinfo{person}{Philip~HS Torr}, {and} \bibinfo{person}{Manmohan
  Chandraker}.} \bibinfo{year}{2017}\natexlab{}.
\newblock \showarticletitle{Desire: Distant future prediction in dynamic scenes
  with interacting agents}. In \bibinfo{booktitle}{\emph{Proceedings of the
  IEEE conference on computer vision and pattern recognition}}.
  \bibinfo{pages}{336--345}.
\newblock


\bibitem[Li et~al\mbox{.}(2020)]%
        {li2020multi}
\bibfield{author}{\bibinfo{person}{Yingying Li}, \bibinfo{person}{Jie Wu},
  \bibinfo{person}{Xue Bai}, \bibinfo{person}{Xipeng Yang},
  \bibinfo{person}{Xiao Tan}, \bibinfo{person}{Guanbin Li},
  \bibinfo{person}{Shilei Wen}, \bibinfo{person}{Hongwu Zhang}, {and}
  \bibinfo{person}{Errui Ding}.} \bibinfo{year}{2020}\natexlab{}.
\newblock \showarticletitle{Multi-granularity tracking with modularlized
  components for unsupervised vehicles anomaly detection}. In
  \bibinfo{booktitle}{\emph{Proceedings of the IEEE/CVF Conference on Computer
  Vision and Pattern Recognition Workshops}}. \bibinfo{pages}{586--587}.
\newblock


\bibitem[Li et~al\mbox{.}(2018)]%
        {li2018diffusion}
\bibfield{author}{\bibinfo{person}{Yaguang Li}, \bibinfo{person}{Rose Yu},
  \bibinfo{person}{Cyrus Shahabi}, {and} \bibinfo{person}{Yan Liu}.}
  \bibinfo{year}{2018}\natexlab{}.
\newblock \showarticletitle{Diffusion Convolutional Recurrent Neural Network:
  Data-Driven Traffic Forecasting}. In \bibinfo{booktitle}{\emph{International
  Conference on Learning Representations}}.
\newblock


\bibitem[Liu et~al\mbox{.}(2022)]%
        {liu2022benchmarking}
\bibfield{author}{\bibinfo{person}{Kay Liu}, \bibinfo{person}{Yingtong Dou},
  \bibinfo{person}{Yue Zhao}, \bibinfo{person}{Xueying Ding},
  \bibinfo{person}{Xiyang Hu}, \bibinfo{person}{Ruitong Zhang},
  \bibinfo{person}{Kaize Ding}, \bibinfo{person}{Canyu Chen},
  \bibinfo{person}{Hao Peng}, \bibinfo{person}{Kai Shu}, {et~al\mbox{.}}}
  \bibinfo{year}{2022}\natexlab{}.
\newblock \showarticletitle{Benchmarking Node Outlier Detection on Graphs}.
\newblock \bibinfo{journal}{\emph{arXiv preprint arXiv:2206.10071}}
  (\bibinfo{year}{2022}).
\newblock


\bibitem[Liu et~al\mbox{.}(2021)]%
        {liu2021anomaly}
\bibfield{author}{\bibinfo{person}{Yixin Liu}, \bibinfo{person}{Shirui Pan},
  \bibinfo{person}{Yu~Guang Wang}, \bibinfo{person}{Fei Xiong},
  \bibinfo{person}{Liang Wang}, \bibinfo{person}{Qingfeng Chen}, {and}
  \bibinfo{person}{Vincent~CS Lee}.} \bibinfo{year}{2021}\natexlab{}.
\newblock \showarticletitle{Anomaly detection in dynamic graphs via
  transformer}.
\newblock \bibinfo{journal}{\emph{IEEE Transactions on Knowledge and Data
  Engineering}} (\bibinfo{year}{2021}).
\newblock


\bibitem[Ma et~al\mbox{.}(2021)]%
        {ma2021comprehensive}
\bibfield{author}{\bibinfo{person}{Xiaoxiao Ma}, \bibinfo{person}{Jia Wu},
  \bibinfo{person}{Shan Xue}, \bibinfo{person}{Jian Yang},
  \bibinfo{person}{Chuan Zhou}, \bibinfo{person}{Quan~Z Sheng},
  \bibinfo{person}{Hui Xiong}, {and} \bibinfo{person}{Leman Akoglu}.}
  \bibinfo{year}{2021}\natexlab{}.
\newblock \showarticletitle{A comprehensive survey on graph anomaly detection
  with deep learning}.
\newblock \bibinfo{journal}{\emph{IEEE Transactions on Knowledge and Data
  Engineering}} (\bibinfo{year}{2021}).
\newblock


\bibitem[Matousek et~al\mbox{.}(2019)]%
        {matousek2019detecting}
\bibfield{author}{\bibinfo{person}{Matthias Matousek},
  \bibinfo{person}{EL-Zohairy Mohamed}, \bibinfo{person}{Frank Kargl},
  \bibinfo{person}{Christoph B{\"o}sch}, {et~al\mbox{.}}}
  \bibinfo{year}{2019}\natexlab{}.
\newblock \showarticletitle{Detecting anomalous driving behavior using neural
  networks}. In \bibinfo{booktitle}{\emph{2019 IEEE Intelligent Vehicles
  Symposium (IV)}}. IEEE, \bibinfo{pages}{2229--2235}.
\newblock


\bibitem[Matousek et~al\mbox{.}(2018)]%
        {matousek2018robust}
\bibfield{author}{\bibinfo{person}{Matthias Matousek}, \bibinfo{person}{Mahmoud
  Yassin}, \bibinfo{person}{Rens van~der Heijden}, \bibinfo{person}{Frank
  Kargl}, {et~al\mbox{.}}} \bibinfo{year}{2018}\natexlab{}.
\newblock \showarticletitle{Robust detection of anomalous driving behavior}. In
  \bibinfo{booktitle}{\emph{2018 IEEE 87th Vehicular Technology Conference (VTC
  Spring)}}. IEEE, \bibinfo{pages}{1--5}.
\newblock


\bibitem[Mohamed et~al\mbox{.}(2020)]%
        {mohamed2020social}
\bibfield{author}{\bibinfo{person}{Abduallah Mohamed}, \bibinfo{person}{Kun
  Qian}, \bibinfo{person}{Mohamed Elhoseiny}, {and} \bibinfo{person}{Christian
  Claudel}.} \bibinfo{year}{2020}\natexlab{}.
\newblock \showarticletitle{Social-stgcnn: A social spatio-temporal graph
  convolutional neural network for human trajectory prediction}. In
  \bibinfo{booktitle}{\emph{Proceedings of the IEEE/CVF Conference on Computer
  Vision and Pattern Recognition}}. \bibinfo{pages}{14424--14432}.
\newblock


\bibitem[Moukafih et~al\mbox{.}(2019)]%
        {moukafih2019aggressive}
\bibfield{author}{\bibinfo{person}{Youness Moukafih}, \bibinfo{person}{Hakim
  Hafidi}, {and} \bibinfo{person}{Mounir Ghogho}.}
  \bibinfo{year}{2019}\natexlab{}.
\newblock \showarticletitle{Aggressive driving detection using deep
  learning-based time series classification}. In \bibinfo{booktitle}{\emph{2019
  IEEE International Symposium on INnovations in Intelligent SysTems and
  Applications (INISTA)}}. IEEE, \bibinfo{pages}{1--5}.
\newblock


\bibitem[Pei et~al\mbox{.}(2022)]%
        {pei2022resgcn}
\bibfield{author}{\bibinfo{person}{Yulong Pei}, \bibinfo{person}{Tianjin
  Huang}, \bibinfo{person}{Werner van Ipenburg}, {and} \bibinfo{person}{Mykola
  Pechenizkiy}.} \bibinfo{year}{2022}\natexlab{}.
\newblock \showarticletitle{ResGCN: attention-based deep residual modeling for
  anomaly detection on attributed networks}.
\newblock \bibinfo{journal}{\emph{Machine Learning}} \bibinfo{volume}{111},
  \bibinfo{number}{2} (\bibinfo{year}{2022}), \bibinfo{pages}{519--541}.
\newblock


\bibitem[Sadeghian et~al\mbox{.}(2019)]%
        {sadeghian2019sophie}
\bibfield{author}{\bibinfo{person}{Amir Sadeghian}, \bibinfo{person}{Vineet
  Kosaraju}, \bibinfo{person}{Ali Sadeghian}, \bibinfo{person}{Noriaki Hirose},
  \bibinfo{person}{Hamid Rezatofighi}, {and} \bibinfo{person}{Silvio
  Savarese}.} \bibinfo{year}{2019}\natexlab{}.
\newblock \showarticletitle{Sophie: An attentive gan for predicting paths
  compliant to social and physical constraints}. In
  \bibinfo{booktitle}{\emph{Proceedings of the IEEE/CVF conference on computer
  vision and pattern recognition}}. \bibinfo{pages}{1349--1358}.
\newblock


\bibitem[Sch{\"o}ller et~al\mbox{.}(2020)]%
        {scholler2020constant}
\bibfield{author}{\bibinfo{person}{Christoph Sch{\"o}ller},
  \bibinfo{person}{Vincent Aravantinos}, \bibinfo{person}{Florian Lay}, {and}
  \bibinfo{person}{Alois Knoll}.} \bibinfo{year}{2020}\natexlab{}.
\newblock \showarticletitle{What the constant velocity model can teach us about
  pedestrian motion prediction}.
\newblock \bibinfo{journal}{\emph{IEEE Robotics and Automation Letters}}
  \bibinfo{volume}{5}, \bibinfo{number}{2} (\bibinfo{year}{2020}),
  \bibinfo{pages}{1696--1703}.
\newblock


\bibitem[Seo et~al\mbox{.}(2018)]%
        {seo2018structured}
\bibfield{author}{\bibinfo{person}{Youngjoo Seo}, \bibinfo{person}{Micha{\"e}l
  Defferrard}, \bibinfo{person}{Pierre Vandergheynst}, {and}
  \bibinfo{person}{Xavier Bresson}.} \bibinfo{year}{2018}\natexlab{}.
\newblock \showarticletitle{Structured sequence modeling with graph
  convolutional recurrent networks}. In \bibinfo{booktitle}{\emph{International
  conference on neural information processing}}. Springer,
  \bibinfo{pages}{362--373}.
\newblock


\bibitem[Stocco et~al\mbox{.}(2022)]%
        {stocco2022thirdeye}
\bibfield{author}{\bibinfo{person}{Andrea Stocco}, \bibinfo{person}{Paulo~J
  Nunes}, \bibinfo{person}{Marcelo D'Amorim}, {and} \bibinfo{person}{Paolo
  Tonella}.} \bibinfo{year}{2022}\natexlab{}.
\newblock \showarticletitle{Thirdeye: Attention maps for safe autonomous
  driving systems}. In \bibinfo{booktitle}{\emph{37th IEEE/ACM International
  Conference on Automated Software Engineering}}. \bibinfo{pages}{1--12}.
\newblock


\bibitem[Stocco et~al\mbox{.}(2020)]%
        {stocco2020misbehaviour}
\bibfield{author}{\bibinfo{person}{Andrea Stocco}, \bibinfo{person}{Michael
  Weiss}, \bibinfo{person}{Marco Calzana}, {and} \bibinfo{person}{Paolo
  Tonella}.} \bibinfo{year}{2020}\natexlab{}.
\newblock \showarticletitle{Misbehaviour prediction for autonomous driving
  systems}. In \bibinfo{booktitle}{\emph{Proceedings of the ACM/IEEE 42nd
  international conference on software engineering}}.
  \bibinfo{pages}{359--371}.
\newblock


\bibitem[Sultani et~al\mbox{.}(2018)]%
        {sultani2018real}
\bibfield{author}{\bibinfo{person}{Waqas Sultani}, \bibinfo{person}{Chen Chen},
  {and} \bibinfo{person}{Mubarak Shah}.} \bibinfo{year}{2018}\natexlab{}.
\newblock \showarticletitle{Real-world anomaly detection in surveillance
  videos}. In \bibinfo{booktitle}{\emph{Proceedings of the IEEE conference on
  computer vision and pattern recognition}}. \bibinfo{pages}{6479--6488}.
\newblock


\bibitem[Sutskever et~al\mbox{.}(2014)]%
        {sutskever2014sequence}
\bibfield{author}{\bibinfo{person}{Ilya Sutskever}, \bibinfo{person}{Oriol
  Vinyals}, {and} \bibinfo{person}{Quoc~V Le}.}
  \bibinfo{year}{2014}\natexlab{}.
\newblock \showarticletitle{Sequence to sequence learning with neural
  networks}.
\newblock \bibinfo{journal}{\emph{Advances in neural information processing
  systems}}  \bibinfo{volume}{27} (\bibinfo{year}{2014}).
\newblock


\bibitem[Tang et~al\mbox{.}(2019)]%
        {tang2019joint}
\bibfield{author}{\bibinfo{person}{Xianfeng Tang}, \bibinfo{person}{Boqing
  Gong}, \bibinfo{person}{Yanwei Yu}, \bibinfo{person}{Huaxiu Yao},
  \bibinfo{person}{Yandong Li}, \bibinfo{person}{Haiyong Xie}, {and}
  \bibinfo{person}{Xiaoyu Wang}.} \bibinfo{year}{2019}\natexlab{}.
\newblock \showarticletitle{Joint modeling of dense and incomplete trajectories
  for citywide traffic volume inference}. In \bibinfo{booktitle}{\emph{The
  World Wide Web Conference}}. \bibinfo{pages}{1806--1817}.
\newblock


\bibitem[Trirat and Lee(2021)]%
        {trirat2021df}
\bibfield{author}{\bibinfo{person}{Patara Trirat} {and}
  \bibinfo{person}{Jae-Gil Lee}.} \bibinfo{year}{2021}\natexlab{}.
\newblock \showarticletitle{Df-tar: a deep fusion network for citywide traffic
  accident risk prediction with dangerous driving behavior}. In
  \bibinfo{booktitle}{\emph{Proceedings of the Web Conference 2021}}.
  \bibinfo{pages}{1146--1156}.
\newblock


\bibitem[Vaswani et~al\mbox{.}(2017)]%
        {vaswani2017attention}
\bibfield{author}{\bibinfo{person}{Ashish Vaswani}, \bibinfo{person}{Noam
  Shazeer}, \bibinfo{person}{Niki Parmar}, \bibinfo{person}{Jakob Uszkoreit},
  \bibinfo{person}{Llion Jones}, \bibinfo{person}{Aidan~N Gomez},
  \bibinfo{person}{{\L}ukasz Kaiser}, {and} \bibinfo{person}{Illia
  Polosukhin}.} \bibinfo{year}{2017}\natexlab{}.
\newblock \showarticletitle{Attention is all you need}.
\newblock \bibinfo{journal}{\emph{Advances in neural information processing
  systems}}  \bibinfo{volume}{30} (\bibinfo{year}{2017}).
\newblock


\bibitem[Veli{\v{c}}kovi{\'c} et~al\mbox{.}(2017)]%
        {velivckovic2017graph}
\bibfield{author}{\bibinfo{person}{Petar Veli{\v{c}}kovi{\'c}},
  \bibinfo{person}{Guillem Cucurull}, \bibinfo{person}{Arantxa Casanova},
  \bibinfo{person}{Adriana Romero}, \bibinfo{person}{Pietro Lio}, {and}
  \bibinfo{person}{Yoshua Bengio}.} \bibinfo{year}{2017}\natexlab{}.
\newblock \showarticletitle{Graph attention networks}.
\newblock \bibinfo{journal}{\emph{arXiv preprint arXiv:1710.10903}}
  (\bibinfo{year}{2017}).
\newblock


\bibitem[Vemula et~al\mbox{.}(2018)]%
        {vemula2018social}
\bibfield{author}{\bibinfo{person}{Anirudh Vemula}, \bibinfo{person}{Katharina
  Muelling}, {and} \bibinfo{person}{Jean Oh}.} \bibinfo{year}{2018}\natexlab{}.
\newblock \showarticletitle{Social attention: Modeling attention in human
  crowds}. In \bibinfo{booktitle}{\emph{2018 IEEE international Conference on
  Robotics and Automation (ICRA)}}. IEEE, \bibinfo{pages}{4601--4607}.
\newblock


\bibitem[Wang and Rajamani(2004)]%
        {1337326}
\bibfield{author}{\bibinfo{person}{Junmin Wang} {and} \bibinfo{person}{R.
  Rajamani}.} \bibinfo{year}{2004}\natexlab{}.
\newblock \showarticletitle{Should adaptive cruise-control systems be designed
  to maintain a constant time gap between vehicles?}
\newblock \bibinfo{journal}{\emph{IEEE Transactions on Vehicular Technology}}
  \bibinfo{volume}{53}, \bibinfo{number}{5} (\bibinfo{year}{2004}),
  \bibinfo{pages}{1480--1490}.
\newblock
\urldef\tempurl%
\url{https://doi.org/10.1109/TVT.2004.832386}
\showDOI{\tempurl}


\bibitem[Wang and Derr(2021)]%
        {wang2021tree}
\bibfield{author}{\bibinfo{person}{Yu Wang} {and} \bibinfo{person}{Tyler
  Derr}.} \bibinfo{year}{2021}\natexlab{}.
\newblock \showarticletitle{Tree decomposed graph neural network}. In
  \bibinfo{booktitle}{\emph{Proceedings of the 30th ACM International
  Conference on Information \& Knowledge Management}}.
  \bibinfo{pages}{2040--2049}.
\newblock


\bibitem[Wang et~al\mbox{.}(2022)]%
        {wang2022graph}
\bibfield{author}{\bibinfo{person}{Yu Wang}, \bibinfo{person}{Wei Jin}, {and}
  \bibinfo{person}{Tyler Derr}.} \bibinfo{year}{2022}\natexlab{}.
\newblock \showarticletitle{Graph neural networks: Self-supervised learning}.
\newblock \bibinfo{journal}{\emph{Graph Neural Networks: Foundations,
  Frontiers, and Applications}} (\bibinfo{year}{2022}),
  \bibinfo{pages}{391--420}.
\newblock


\bibitem[Wiederer et~al\mbox{.}(2022)]%
        {wiederer2022anomaly}
\bibfield{author}{\bibinfo{person}{Julian Wiederer}, \bibinfo{person}{Arij
  Bouazizi}, \bibinfo{person}{Marco Troina}, \bibinfo{person}{Ulrich Kressel},
  {and} \bibinfo{person}{Vasileios Belagiannis}.}
  \bibinfo{year}{2022}\natexlab{}.
\newblock \showarticletitle{Anomaly Detection in Multi-Agent Trajectories for
  Automated Driving}. In \bibinfo{booktitle}{\emph{Conference on Robot
  Learning}}. PMLR, \bibinfo{pages}{1223--1233}.
\newblock


\bibitem[Wu et~al\mbox{.}(2021)]%
        {wu2021box}
\bibfield{author}{\bibinfo{person}{Jie Wu}, \bibinfo{person}{Xionghui Wang},
  \bibinfo{person}{Xuefeng Xiao}, {and} \bibinfo{person}{Yitong Wang}.}
  \bibinfo{year}{2021}\natexlab{}.
\newblock \showarticletitle{Box-level tube tracking and refinement for vehicles
  anomaly detection}. In \bibinfo{booktitle}{\emph{Proceedings of the IEEE/CVF
  Conference on Computer Vision and Pattern Recognition}}.
  \bibinfo{pages}{4112--4118}.
\newblock


\bibitem[Xu et~al\mbox{.}(2022)]%
        {xu2022contrastive}
\bibfield{author}{\bibinfo{person}{Zhiming Xu}, \bibinfo{person}{Xiao Huang},
  \bibinfo{person}{Yue Zhao}, \bibinfo{person}{Yushun Dong}, {and}
  \bibinfo{person}{Jundong Li}.} \bibinfo{year}{2022}\natexlab{}.
\newblock \showarticletitle{Contrastive Attributed Network Anomaly Detection
  with Data Augmentation}. In \bibinfo{booktitle}{\emph{Pacific-Asia Conference
  on Knowledge Discovery and Data Mining}}. Springer,
  \bibinfo{pages}{444--457}.
\newblock


\bibitem[Yu et~al\mbox{.}(2018)]%
        {yu2018netwalk}
\bibfield{author}{\bibinfo{person}{Wenchao Yu}, \bibinfo{person}{Wei Cheng},
  \bibinfo{person}{Charu~C Aggarwal}, \bibinfo{person}{Kai Zhang},
  \bibinfo{person}{Haifeng Chen}, {and} \bibinfo{person}{Wei Wang}.}
  \bibinfo{year}{2018}\natexlab{}.
\newblock \showarticletitle{Netwalk: A flexible deep embedding approach for
  anomaly detection in dynamic networks}. In
  \bibinfo{booktitle}{\emph{Proceedings of the 24th ACM SIGKDD international
  conference on knowledge discovery \& data mining}}.
  \bibinfo{pages}{2672--2681}.
\newblock


\bibitem[Zheng et~al\mbox{.}(2019)]%
        {zheng2019addgraph}
\bibfield{author}{\bibinfo{person}{Li Zheng}, \bibinfo{person}{Zhenpeng Li},
  \bibinfo{person}{Jian Li}, \bibinfo{person}{Zhao Li}, {and}
  \bibinfo{person}{Jun Gao}.} \bibinfo{year}{2019}\natexlab{}.
\newblock \showarticletitle{AddGraph: Anomaly Detection in Dynamic Graph Using
  Attention-based Temporal GCN.}. In \bibinfo{booktitle}{\emph{IJCAI}}.
  \bibinfo{pages}{4419--4425}.
\newblock


\end{thebibliography}

\appendix
\section{Data details}
We describe the detailed settings for simulation data, then provide the data statistics for both simulation and highD data.
\begin{itemize}
    \item Normal traffic.  A standard car following model, \textit{Modified General Motors} ~\cite{ahmed1999modeling} is used, which has been demonstrated to correlate well with field traffic data. The desired speed of the vehicles follows a typical distribution found in real-world traffic, with the majority between 65-80 mph, only 5\% above 85 mph and 5\% below 60 mph. Recordings of varying traffic demands from 500 to 1600 vehicles/lane/hour are included, covering both free flow and congested conditions.
    
    \item Speeding scenario.  We set 70\% of the vehicles drive at a desired speed of 65 mph, and 30\% above 85 mph. The traffic demand is 500 veh/l/hr.

    \item Slow scenario. We set 98\% of the vehicles drive at a desired speed of 65 mph, and 2\% below 50 mph. The traffic demand is 500 veh/l/hr.
    
    \item Tailgating scenario. We set 46.80\% cars to follow a \textit{Constant Time Gap} model and only a proportion of them could be tailgaters according to traffic conditions. The traffic demand is 500 veh/l/hr.
    
    \item Stalled car scenario. We randomly select 15 cars, each stopping for 5 minutes. The traffic demand is 1500 veh/l/hr.
    
    \item Comprehensive scenario. We set 89\% of the vehicles drive at a desired speed of 65 mph, 10\% at 85 mph, and 1\% at 50 mph. In addition, we set 46.80\% cars following \textit{Constant Time Gap} model and only a proportion of them could be tailgaters according to traffic condition. We set 2 cars to each stall for 3 min. The traffic demand is 1000 veh/l/hr.
\end{itemize}

Cars are labeled anomaly only when it is actually behaving anomalously (e.g., when a car with desired speed of 85 mph can only drive at 65 mph because of traffic conditions, it is not an anomaly at the corresponding time).

Table~\ref{table:count} shows the statistics of car count, and the total trajectory count when divided into 10-15s time windows with 1s stride.
We note that in simulation data, the traffic flow in stalled car and comprehensive scenarios are higher, because we want to check if we can detect the source anomaly car even if the stalled car causes upstream congestion.

\begin{table}
\centering
\caption{Dataset statistics}
\label{table:count}
\begin{tabular}{l|l|ll} 
\toprule
                            &               & Car count & Trajectory count  \\ 
\midrule
\multirow{6}{*}{Simulation} & training      & 7,886     & 920,311           \\
                            & comprehensive & 1,281     & 147,254           \\
                            & slow          & 636       & 75,809            \\
                            & speeding      & 583       & 65,685            \\
                            & tailgating    & 645       & 78,600            \\
                            & stalled       & 1,922     & 263,370           \\ 
\midrule
\multirow{2}{*}{HighD}      & training      & 42,106    & 459,187           \\
                            & Testing       & 1,795     & 21,840            \\
\bottomrule
\end{tabular}
\end{table}

\section{Description of Metrics}
In this section, we describe the evaluation metrics.

\textit{i}) \textit{area under the receiver operating characteristic curve} (ROC-AUC) score. The ROC curve plots the true positive rate (TPR) against the false positive rate (FPR), and the ROC-AUC score calculates the area under the ROC curve. An ROC-AUC score of 0.5 means the model is not able to discriminate anomalies, and an ROC-AUC score of 1 means perfect anomaly detection. 

\textit{ii}) Average precision, which summarizes the precision-recall curve into a single value, and is calculated as the weighted mean of previsions achieved at each threshold, the weight being the increase in recall from the previous threshold. 

\textit{iii}) Precision@k, which calculates the percentage of true anomaly among the top k samples scored by the models to be anomalies.

We note that for vehicle detection problem, since the anomaly rate is only 3\%-5\%, severe data imbalance issue exists. It is shown in \cite{jeni2013facing} that in heavily imbalanced datasets, metrics like F1 score downgrades exponentially with data skewness, thus we do not include F1 metrics. Meanwhile, ROC-AUC score is less influenced by data imbalance, but with a caveat that model with distinctive performance could have very similar ROC-AUC score. Precision@k shows how well we rank anomalies over normal samples, and works well on imbalanced datasets where our focus is on the relatively rarely occurring anomalies.

\section{Aggregator} \label{sec:aggregator}
\edit{We briefly describe the choice of loss aggregation over time and vehicles in section~\ref{sec:agg}. Averaging the losses is more conservative than maximizing. For vehicle detection, since many abnormal behaviors persist for relatively long time periods over several seconds, we choose to average over time. Maximization results in  around 0.1 decrease in average precision. For scene detection, we choose maximization to avoid abnormal vehicles being averaged out by normal vehicles. Averaging results in around 0.06 decrease in average precision.}

\section{Robustness analysis} \label{sec:clean_training}
\edit{We briefly explore the model robustness to anomalies in training data. As described in section~\ref{sec:dataset}, our training data contains a small portion of anomalies, to simulate the real-world situation of not having a perfectly clean normal dataset. To evaluate the influence of training anomalies, we compare the performance with model trained on clean training data. Table~\ref{table:clean_training} shows the result. We can see that the results are similar, with differences in scores all less than 0.05. There is also no clear trend of one dataset better than the other. The result shows the model is robust to small  number of anomalies in the training data.}

\begin{table}
\centering
\caption{\edit{Influence of anomalies in the training data. The performance of two training sets are comparable, with differences in scores all less than 0.05.}}
\label{table:clean_training}
\begin{tabular}{l|l|llll} 
\toprule
                                                                             & \begin{tabular}[c]{@{}l@{}}Training\\~data\end{tabular} & \multicolumn{1}{c}{Pre@100} & \multicolumn{1}{c}{Pre@200} & \multicolumn{1}{c}{Avg Pre} & \multicolumn{1}{c}{AUC}  \\ 
\midrule
\multirow{2}{*}{\begin{tabular}[c]{@{}l@{}}vehicle \\detection\end{tabular}} & Clean                                                   & 0.79                        & 0.815                       & 0.336                       & 0.888                    \\
                                                                             & Polluted                                                & 0.82                        & 0.775                       & 0.381                       & 0.900                    \\ 
\midrule
\multirow{2}{*}{\begin{tabular}[c]{@{}l@{}}scene\\detection\end{tabular}}    & Clean                                                   & 0.97                        & 0.965                       & 0.852                       & 0.834                    \\
                                                                             & Polluted                                                & 0.93                        & 0.950                       & 0.859                       & 0.841                    \\
\bottomrule
\end{tabular}
\end{table}

\section{HighD anomaly description} \label{sec:highD_more}
\begin{figure}
\centering

\begin{subfigure}[t]{0.85\columnwidth}
    \includegraphics[width=\linewidth]{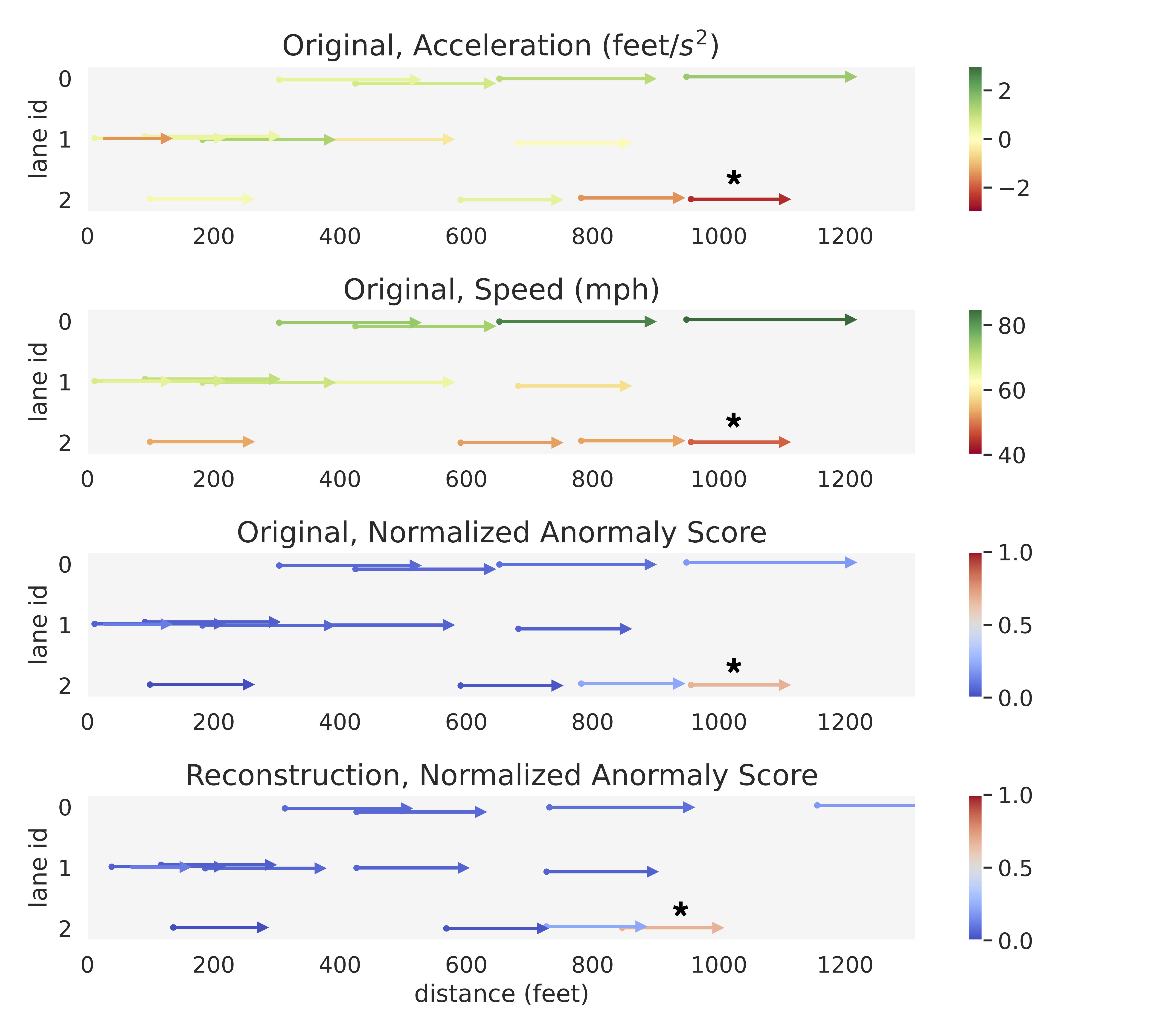}
    \caption{\normalfont{Vehicle with 3rd largest anomaly score. The abnormal vehicle (marked $\ast$) is having dramatic deceleration and drives significantly slower than other cars in its lane. }}
    \Description{Four rectangular figures containing vehicle trajectories with x-axis as distance and y-axis as lane id. There are multiple lines on the figure, each of which denotes the trajectory of a single vehicle, with a dot denoting the starting point and a triangle denoting the endpoint. The lines have different color bars to represent different features.
    
    For the first top figure, the color represents vehicle acceleration where the color bar is from red to yellow to green and represents -2 to 0 to 2. There is one extremely red line with starting point at lane 2 and end point at lane 2. This line is marked by \*. The rest lines have relatively more yellow or green colors. For the second top figure, the color represents vehicle speed where the color bar is from red to yellow to green and represents 40 to 60 to 80. All lines have a similar pattern with the first top figure but with different colors. For the third figure, the color represents each vehicle's original normalized anomaly score where the color bar is from blue to white to red and represents 0 to 0.5 to 1. The red line mentioned in the first top figure also has red color here while the rest lines are all blue. For the fourth figure, the color represents each vehicle's reconstructed normalized anomaly score where the color bar is from blue to white to red and represents 0 to 0.5 to 1. The red line mentioned in the first top figure also has red color here while the rest lines are all blue.}
    \label{fig:highD_abn3}
\end{subfigure}

\quad
\begin{subfigure}[t]{0.85\columnwidth}
    \includegraphics[width=\linewidth]{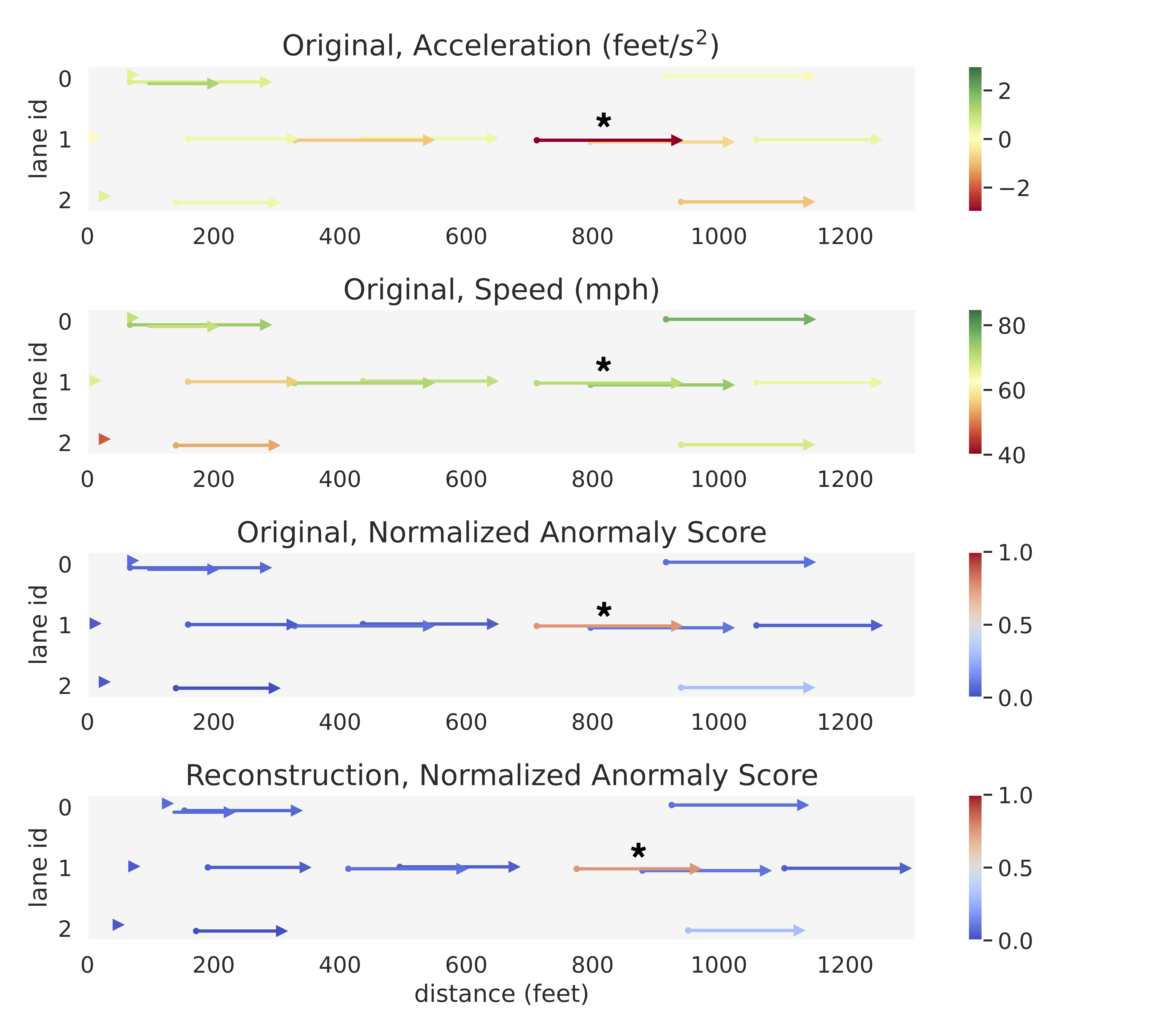}
    \caption{\normalfont{Vehicle with 4th largest anomaly score. The abnormal vehicle (marked $\ast$) is having dramatic deceleration.} }
    \Description{Four rectangular figures containing vehicle trajectories with x-axis as distance and y-axis as lane id. There are multiple lines on the figure, each of which denotes the trajectory of a single vehicle, with a dot denoting the starting point and a triangle denoting the endpoint. The lines have different color bars to represent different features.
    
    For the first top figure, the color represents vehicle acceleration where the color bar is from red to yellow to green and represents -2 to 0 to 2. There is one extremely red line with starting point at lane 1 and end point at lane 1. This line is marked by \*. The rest lines have relatively more yellow or green colors. For the second top figure, the color represents vehicle speed where the color bar is from red to yellow to green and represents 40 to 60 to 80. All lines have a similar pattern to the first top figure but with different colors. For the third figure, the color represents each vehicle's original normalized anomaly score where the color bar is from blue to white to red and represents 0 to 0.5 to 1. The red line mentioned in the first top figure also has red color here while the rest lines are all blue. For the fourth figure, the color represents each vehicle's reconstructed normalized anomaly score where the color bar is from blue to white to red and represents 0 to 0.5 to 1. The red line mentioned in the first top figure also has red color here while the rest lines are all blue.}
    \label{fig:highD_abn4}
\end{subfigure}

\caption{Additional qualitative study of real-world HighD traffic data (Part1)}
\label{fig:highD2}
\end{figure}

\begin{figure}
\centering

\begin{subfigure}[t]{0.85\columnwidth}
    \includegraphics[width=\linewidth]{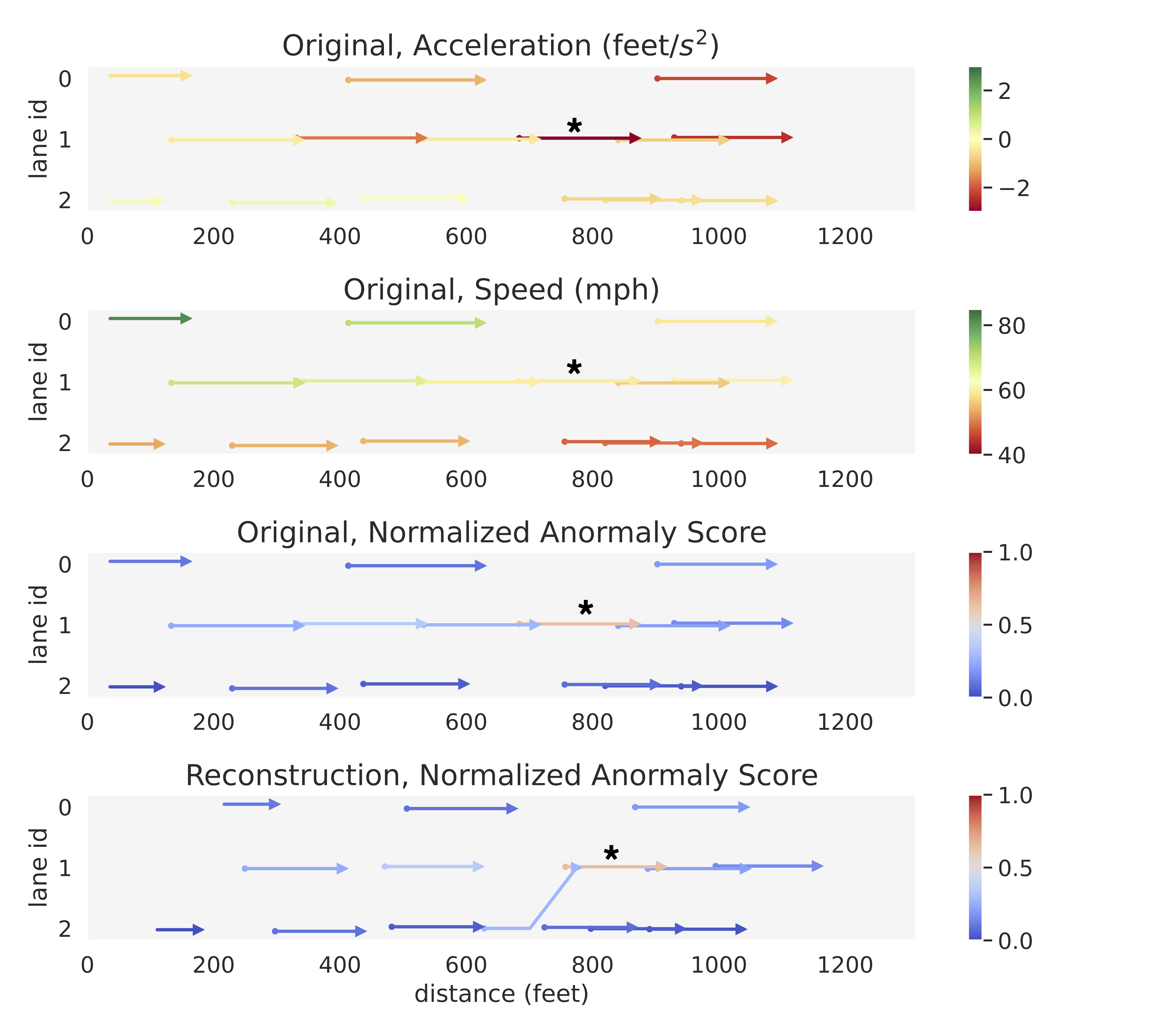}
    \caption{\normalfont{ Vehicle with 5th largest anomaly score. The abnormal vehicle (marked $\ast$) is having dramatic deceleration.}}
    \Description{Four rectangular figures containing vehicle trajectories with x-axis as distance and y-axis as lane id. There are multiple lines on the figure, each of which denotes the trajectory of a single vehicle, with a dot denoting the starting point and a triangle denoting the endpoint. The lines has different colors to represent different features.
    
    For the first top figure, the color represents vehicle acceleration where the color bar is from red to yellow to green and represents -2 to 0 to 2. There is one extremely red line with starting point at lane 2 and end point at lane 2. This line is marked by \*. The rest lines have relatively more yellow or green colors. For the second top figure, the color represents vehicle speed where the color bar is from red to yellow to green and represents 40 to 60 to 80. All lines have a similar pattern with the first top figure but with different colors. For the third figure, the color represents each vehicle's original normalized anomaly score where the color bar is from blue to white to red and represents 0 to 0.5 to 1. The red line mentioned in the first top figure also has red color here while the rest lines are all blue. For the fourth figure, the color represents each vehicle's reconstructed normalized anomaly score where the color bar is from blue to white to red and represents 0 to 0.5 to 1. The red line mentioned in the first top figure also has red color here while the rest lines are all blue.}
    \label{fig:highD_abn5}
\end{subfigure}

\begin{subfigure}[t]{0.9\columnwidth}
    \includegraphics[width=\linewidth]{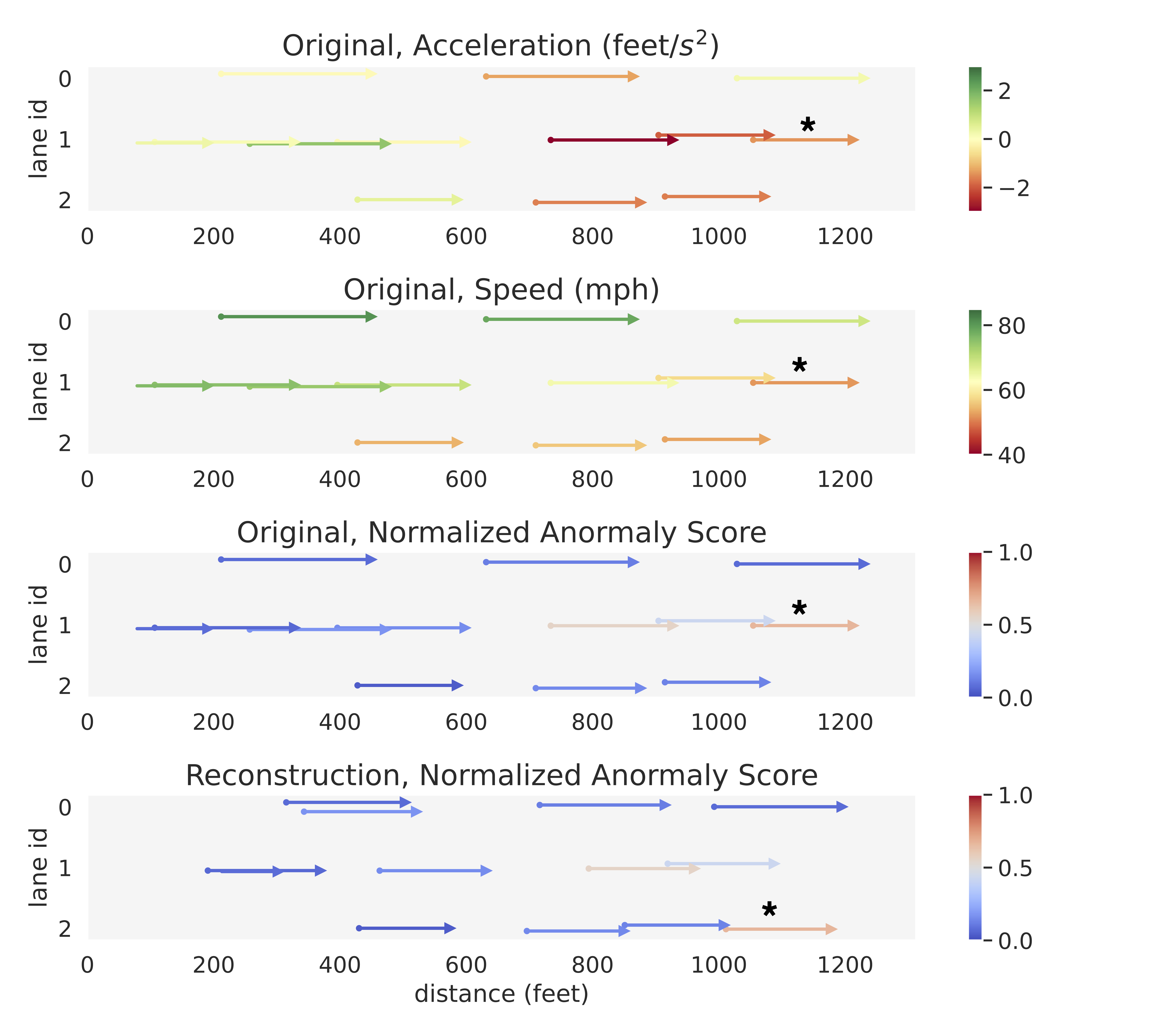}
    \caption{\normalfont{Vehicle with 6th largest anomaly score. The abnormal vehicle (marked $\ast$) is having dramatic deceleration, while also driving too slow with respect to the lane it is in, therefore reconstructed to rightmost lane which has smaller typical speed.}}
    \Description{Four rectangular figures containing vehicle trajectories with x-axis as distance and y-axis as lane id. There are multiple lines on the figure, each of which denotes the trajectory of a single vehicle, with a dot denoting the starting point and a triangle denoting the endpoint. The lines has different colors to represent different features.
    
    For the first top figure, the color represents vehicle acceleration where the color bar is from red to yellow to green and represents -2 to 0 to 2. There is one extremely red line with starting point at lane 2 and end point at lane 2. This line is marked by \*. The rest lines have relatively more yellow or green colors. For the second top figure, the color represents vehicle speed where the color bar is from red to yellow to green and represents 40 to 60 to 80. All lines have a similar pattern with the first top figure but with different colors. For the third figure, the color represents each vehicle's original normalized anomaly score where the color bar is from blue to white to red and represents 0 to 0.5 to 1. The red line mentioned in the first top figure also has red color here while the rest lines are all blue. For the fourth figure, the color represents each vehicle's reconstructed normalized anomaly score where the color bar is from blue to white to red and represents 0 to 0.5 to 1. The red line mentioned in the first top figure also has red color here while the rest lines are all blue.}
    \label{fig:highD_abn6}
\end{subfigure}


\caption{Additional qualitative study of real-world HighD traffic data (Part 2)}
\label{fig:highD3}
\end{figure}

\edit{In this section, we provide more examples of the qualitative results for top ranked anomalies in the HighD real-world data. We note that during ranking, there might be samples with overlapping time windows, and we eliminate the repetitions for diversity. The results are shown in Fig~\ref{fig:highD2} and Fig~\ref{fig:highD3}. Out of the top anomalies, most are because of drastic deceleration, as well as a significant difference in speed relative to the speed of the surrounding cars, or to the typical speed of the corresponding lane it is in.}


\end{document}